\documentclass[manuscript]{acmart}

\usepackage{url}
\usepackage{float}
\usepackage[nolist]{acronym}
\usepackage{subcaption}
\usepackage{makecell}

\acrodef{FI}[FI]{feature importance}
\acrodef{ML}[ML]{Machine Learning}
\acrodef{XAI}[XAI]{eXplainable Artificial Intelligence}
\acrodef{LIME}[LIME]{Linear Interpretable Model-Agnostic Explanations}
\acrodef{SHAP}[SHAP]{SHapley Additive exPlanations}
\acrodef{LRP}[LRP]{Layer-wise Relevance Propagation}
\acrodef{IntGrad}[IntGrad]{Integrated Gradients}
\acrodef{Grad-CAM}[Grad-CAM]{Gradient-weighted Class Activation Mapping}

\AtBeginDocument{%
  \providecommand\BibTeX{{%
    \normalfont B\kern-0.5em{\scshape i\kern-0.25em b}\kern-0.8em\TeX}}}

\acmConference[ACM FAccT '24]{Make sure to enter the correct
  conference title from your rights confirmation emai}{June 03--06,
  2024}{Rio de Janeiro, Brazil}

\copyrightyear{2024} 
\acmYear{2024} 
\setcopyright{rightsretained} 
\acmConference[FAccT '24]{The 2024 ACM Conference on Fairness, Accountability, and Transparency}{June 3--6, 2024}{Rio de Janeiro, Brazil}
\acmBooktitle{The 2024 ACM Conference on Fairness, Accountability, and Transparency (FAccT '24), June 3--6, 2024, Rio de Janeiro, Brazil}\acmDOI{10.1145/3630106.3658537}
\acmISBN{979-8-4007-0450-5/24/06}
\begin{document}

\title[Classification Metrics for Image Explanations]{Classification Metrics for Image Explanations}
\subtitle{Towards Building Reliable XAI-Evaluations}

\author{Benjamin Fresz}
\authornote{Both authors contributed equally to this research.}
\affiliation{%
  \institution{Fraunhofer Institute for Manufacturing Engineering and Automation IPA}
  \streetaddress{Nobelstrasse 12}
  \city{Stuttgart}
  \country{Germany,}
  \postcode{70569}}
\affiliation{%
  \institution{Institute of Industrial Manufacturing and Management IFF, University of Stuttgart}
  \city{Stuttgart}
  \country{Germany}}
\email{benjamin.fresz@ipa.fraunhofer.de}
\orcid{0009-0002-7463-8907}

\author{Lena Loercher}
\authornotemark[1]
\affiliation{%
  \institution{Fraunhofer Institute for Manufacturing Engineering and Automation IPA}
  \streetaddress{Nobelstrasse 12}
  \city{Stuttgart}
  \country{Germany}
  \postcode{70569}}
\email{lena.loercher@ipa.fraunhofer.de}
\orcid{0000-0002-0109-0226}

\author{Marco F. Huber}
\affiliation{%
  \institution{Fraunhofer Institute for Manufacturing Engineering and Automation IPA}
  \streetaddress{Nobelstrasse 12}
  \city{Stuttgart}
  \country{Germany,}
  \postcode{70569}}
\affiliation{%
  \institution{Institute of Industrial Manufacturing and Management IFF, University of Stuttgart}
  \city{Stuttgart}
  \country{Germany}}
\email{marco.huber@ipa.fraunhofer.de}
\orcid{0000-0002-8250-2092}


\begin{abstract}
    Decision processes of computer vision models---especially deep neural networks---are opaque in nature, meaning that these decisions cannot be understood by humans. Thus, over the last years, many methods to provide human-understandable explanations have been proposed. For image classification, the most common group are saliency methods, which provide (super-)pixelwise feature attribution scores for input images. But their evaluation still poses a problem, as their results cannot be simply compared to the unknown ground truth. To overcome this, a slew of different proxy metrics have been defined, which are---as the explainability methods themselves---often built on intuition and thus, are possibly unreliable.
    In this paper, new evaluation metrics for saliency methods are developed and common saliency methods are benchmarked on ImageNet. In addition, a scheme for reliability evaluation of such metrics is proposed that is based on concepts from psychometric testing. The used code can be found at this \href{https://github.com/lelo204/ClassificationMetricsForImageExplanations}{link}.
\end{abstract}


\received{22 January 2024}

\maketitle

\section{Introduction}
In recent years, \ac{XAI} has gained significant attention as a means to address the black-box nature of many \ac{ML} models.
\ac{XAI} methods aim to provide transparency and interpretability, allowing users to understand the decision-making process of \ac{ML} models.
While various \ac{XAI} techniques have been developed, their evaluation remains challenging, particularly in computer vision tasks. 
A common approach of explaining image classification and object detection decisions are so-called \emph{saliency maps} that highlight image regions deemed particularly important for the prediction.
The evaluation of such methods for image classification is essential to assess their effectiveness and compare different approaches.
However, this is still an open problem despite various approaches to assess the properties of the saliency method, mainly due to the subjective nature of evaluations \cite{rong2023humancentered}, the fallibility of user studies \cite{chromik2021illusionExDepth}, and the different concepts used to evaluate such metrics \cite{watson2020conceptualChallenges}. 
It is particularly difficult to compare and assess saliency explanations beyond anecdotal evidence, as by definition they only provide local explanations, i.e., explanations for individual data points.
A remedy for this can be ways to evaluate local explanations over entire datasets as in \cite{arias2022focus}, resulting in a global assessment of explanation properties.

For \ac{XAI} methods in general, the lack of ground-truth explanations complicates their robust assessment, sometimes attempted to be solved via creating specific datasets with ground-truth explanations \cite{amiri2020groundTruthExplanations, arras2022groundTruthDataset}.
Additional to using such datasets, concepts from other disciplines with similar problems---lack of ground truth and tests with possibly differing underlying concepts---such as psychometric testing can be used \cite{tomsett2019sanityformetrics}. 

The aim of this work is to further develop the ideas of \citet{arias2022focus, AriasDuart2023confusionMatrix}, which provide a set of metrics for saliency methods.
This set is extended to a comprehensive list of metrics that mimic common metrics for classification evaluation based on the definition of correct and incorrect \ac{FI} in images.
Additionally, it is shown how the reliability (as part of validity) of the proposed metrics can be assessed, based on \cite{tomsett2019sanityformetrics}.
As such, our contributions are:
\begin{itemize}
    \item We extend the list of saliency metrics. While \cite{AriasDuart2023confusionMatrix} introduces some of them, others are overlooked. The additional metrics in particular provide interesting additional information, as shown in Section~\ref{sec:results}.
    \item We show how such metrics can be assessed regarding reliability (as a precursor to validity) in order to test them for their practical use.
    \item With the full set of saliency metrics and by adding B-cos networks \cite{böhle2023bcos} and the popular SHAP \cite{lundberg2017shap}, we provide a more complete benchmark of \ac{XAI} methods.
    \item We provide an in-depth discussion of the saliency metrics and show which properties of \ac{XAI} methods (as given by \citet{Nauta2023XAIevalsurvey}) they address.    
\end{itemize}
In the following chapter, the Focus score of \citet{arias2022focus} is briefly introduced, the saliency methods used in this paper and further works related to \ac{XAI}-evaluation are discussed, with the proposed methodology being presented in Section~\ref{sec:methodology}, including the definition of (in)correct \ac{FI} and the psychometric evaluation approach used here. 
The experiment setup and specifically created datasets are described in Section~\ref{sec:experiments}, followed by a selection of the results for the proposed metrics in Section~\ref{sec:results}, accompanied by a discussion of the limitations of the approach in Section~\ref{sec:discussion}. The paper is closed with a summary in Section~\ref{sec:summary}.


\section{Related Work} \label{sec:related-work}

This work builds on the idea of \citet{arias2022focus}, where a metric for evaluating \ac{XAI} methods is proposed. Since there is no ground truth for individual image pixels as to which class they can be assigned to, the authors have proposed a different way of evaluating explanations of saliency methods: they create mosaics from four images of various classes and assume that the evidence towards a class is more prevalent in images labelled with that class. The explanations, which are provided as \ac{FI} by the examined \ac{XAI} methods, are then evaluated by comparing positive feature attribution on images belonging to the correct class with positive feature attribution on the entire mosaic. Positive feature attribution here refers to the summed up feature attribution values of each pixel in the respective part of the mosaic.

In Section~\ref{ssec:saliency-methods} the saliency methods considered in the original paper \cite{arias2022focus} as well as some additional methods examined in this paper are briefly presented. Section~\ref{ssec:xai-eval} addresses the state of the art with regard to \ac{XAI} evaluation.


\subsection{Saliency Methods}\label{ssec:saliency-methods}

The saliency methods for which the Focus score was determined and analyzed in \cite{arias2022focus} will be described shortly in the following. \ac{LIME} was one of the first methods to provide model-agnostic explanations in the form of feature attribution. The feature attribution is calculated by sampling around a data-point and fitting a simpler linear model to the weighted samples \cite{ribeiro2016lime}. For images, \ac{LIME} can create class-specific explanations by highlighting image regions---so-called superpixels---that are deemed especially relevant for the target class. \ac{LRP}, on the other hand, uses first-order Taylor expansions for local renormalization layers to generate saliency maps \cite{binder2016lrp}. \ac{IntGrad}, which was proposed in \cite{sundararajan2017intgrad}, calculates the feature importance of an image by forming the gradient of the model output with respect to the model input and integrating this gradient over a baseline image (``neutral'' input, e.g., grey image).
\ac{Grad-CAM} produces class-specific saliency maps by computing gradient information in the last convolutional layer of a neural network \cite{selvaraju2016gradcam}. A modification of \ac{Grad-CAM}, namely \ac{Grad-CAM}++, uses a weighted combination of the positive partial derivatives in the last convolutional layer, improving the performance of \ac{Grad-CAM} for multiple objects of the same class in a single image and object localization \cite{chattopadhyay2017gradcampp}. Lastly, SmoothGrad describes the exchange of the often noisy gradient-based explanations by a weighted local average, thus possibly improving the visual quality and informativeness \cite{smilkov2017smoothgrad}.

In this work, the range of examined saliency methods is extended to also include \ac{SHAP} and B-cos. \ac{SHAP} is not an explanation method itself but a unifying framework for feature attribution methods, especially Shapley Regression, Shapley Sampling, Quantitative Input Influence Feature Attributions, \ac{LIME}, DeepLIFT, and \ac{LRP} \cite{lundberg2017shap}. The game-theoretic interpretation of these methods, which are used to approximate Shapley Values (given certain hyperparameter choices), provides the possibility of receiving feature attributions with three desired criteria: Local accuracy, missingness, and consistency. KernelSHAP was chosen to approximate Shapley Values here, due to its comparatively low runtime. In contrast to all the post-hoc explanation methods described before, B-cos networks \cite{böhle2023bcos} generate model-inherent saliency maps by changing the activation functions of neural networks. This forces the network weights to align with the network-input and requires the networks to be trained with the B-cos transform as activation functions.

\subsection{XAI Evaluation}\label{ssec:xai-eval}

In recent years, several approaches have been proposed for evaluating \ac{XAI} methods. An overview of the methods published by the end of 2020 can be found in \cite{Nauta2023XAIevalsurvey}. The authors list twelve properties of \ac{XAI} methods that can be tested, often with various automated checks assessing (part of) a specific property.
They group these properties into user-, presentation-, and content-properties, with six of them belonging to the last class and three to each of the previous ones.
The content-properties are the most likely ones to be objectively measurable, although the authors of this paper expect that single metrics will most likely only assess a small subset of the available properties at once, as most of them are quite disjoint in their interpretation, e.g., covariate complexity---denoting how complex the (interactions of) features in the explanation are---and consistency---denoting how deterministic and implementation-invariant an explanation is---probably require quite different assessment methods.
The metrics in this paper are therefore limited to assessing two of these properties: The \emph{contrastivity} of saliency explanations, denoting how strongly an explanation discriminates between different outcomes of the \ac{ML} model. An explanation that does not discriminate well would probably highlight general information such as edges in the mosaic images, thus resulting in bad saliency metrics.
Additionally, with the assumption fulfilled that the used models are able to distinguish between the relevant classes and evidence can mainly be found within images of the corresponding class, the \emph{correctness} of saliency methods can be assessed (as described in the beginning of this chapter).

The explanation type and thus, the evaluation usually depends on the type of input data for which predictions or models need to be explained. Because of this, available toolboxes are limited to certain data types while still providing multiple evaluation metrics, e.g., \cite{agarwal2023openxai} for tabular explanations and \cite{hedström2023quantus} for image explanations. \citet{doshivelez2017rigorousXAI} proposed three different stages of \ac{XAI} evaluation, each with increasing effort and cost: functionally-grounded, human-grounded, and application-grounded. As part of the functionally-grounded evaluation and thus, early on in the development and implementation of \ac{XAI} methods, metrics such as the one presented here can be used.

User studies are often viewed as the gold-standard of \ac{XAI} evaluation, although their results have to be taken with caution as users tend to overestimate their understanding of the \ac{ML} model \cite{vanderwaa2021ruleVsExampleUserstudy, chromik2021illusionExDepth}, which distorts the study results. In addition to user studies and metrics that can be evaluated on a specific use-case, more general documentation approaches have been suggested, for instance Explanation Fact Sheets \cite{Sokol2020ExplFactSheets}, which contain information on relevant aspects of \ac{XAI} methods. A similar approach, although more anecdotal and use-case specific, can be found in \cite{boggust2023saliencycards}, which aims at providing a standardized format to assess and discuss trade-offs when evaluating saliency methods.

In the absence of established ways to compare \ac{XAI} methods on a non-task-specific basis, so-called \emph{sanity checks} can be used. These can test saliency methods for image classification for desiderata such as model-invariance and input-invariance \cite{adebayo2018sanityChecks, kindermans2017unreliabilitySaliency}. Even though a successful check does not provide enough information to fully trust a model, an unsuccessful one does show problematic behavior. Such sanity checks can also be formulated for object detection models \cite{padmanabhan2023sanityObjectDetectors}, although the idea of general sanity checks and the ones which are not task-specific can be criticized due to possibly introducing a selection bias 
\cite{yona2021sanityChecksRevisit}. Other sanity checks involve the creation of ``ground-truth'' saliency maps that are compared with the generated explanations \cite{kim2021sanitySimulations}.

\citet{rao2022DiPart} propose a metric similar to Focus \cite{arias2022focus}, which also only uses positive feature attributions, but limit all classes to appear in the mosaic at most once.
The authors guarantee the basic assumption of class-specific features occurring exclusively in the target class by constraining the classifier to use the information from one part of the mosaic only.
This is done by building one separate model head for every image in the mosaic.
This ensures that no visual information between images is exchanged, limiting the classifier in its decision to rely on single images. The authors then define their metric based on whether the saliency methods still highlight other parts of the mosaic. Moreover, they evaluate the mosaics visually by humans via a systematic assessment approach that entails clustering them with their previously defined scores.

In \cite{vandersmissen2023saliencyMetricBenchmark} a benchmark of common saliency methods and evaluation metrics is provided. The work concludes that the evaluation results are inconclusive and the metrics in part contradict each other.

Finally, \citet{tomsett2019sanityformetrics} investigate \ac{XAI} evaluation metrics and present an approach from psychometric testing to assess them. Apart from \cite{tomsett2019sanityformetrics}, however, there is little research that addresses the topic of \ac{XAI} metrics evaluation.

\section{Methodology} \label{sec:methodology}
This section extends the Focus score for evaluating \ac{XAI} methods from \citet{arias2022focus} by incorporating negative \ac{FI}. First, the construction of mosaics is explained, followed by the definition of true positives and negatives, and false positives and negatives with respect to \ac{FI} in the mosaics. These are then used to define additional evaluation metrics. In the second part, an approach from psychometric testing is introduced to examine the suitability of these metrics for evaluating \ac{XAI} methods.

\subsection{Proposed Metrics}
To calculate the saliency metrics, so-called \emph{mosaics} are used.
They consist of a $2 \times 2$ grid of images of different classes from the original dataset.
The idea behind them is that---given a model that is able to distinguish between the relevant classes---\ac{FI} for a given class $C$ should be attributed to the part of the mosaic that belongs to class $C$ (as denoted by the labels in the original dataset).
This then allows to calculated metrics akin to classical metrics for classification tasks, as described in the following.

\subsubsection{Mosaics}
\begin{figure*}[t]
    \includegraphics[width=\textwidth]{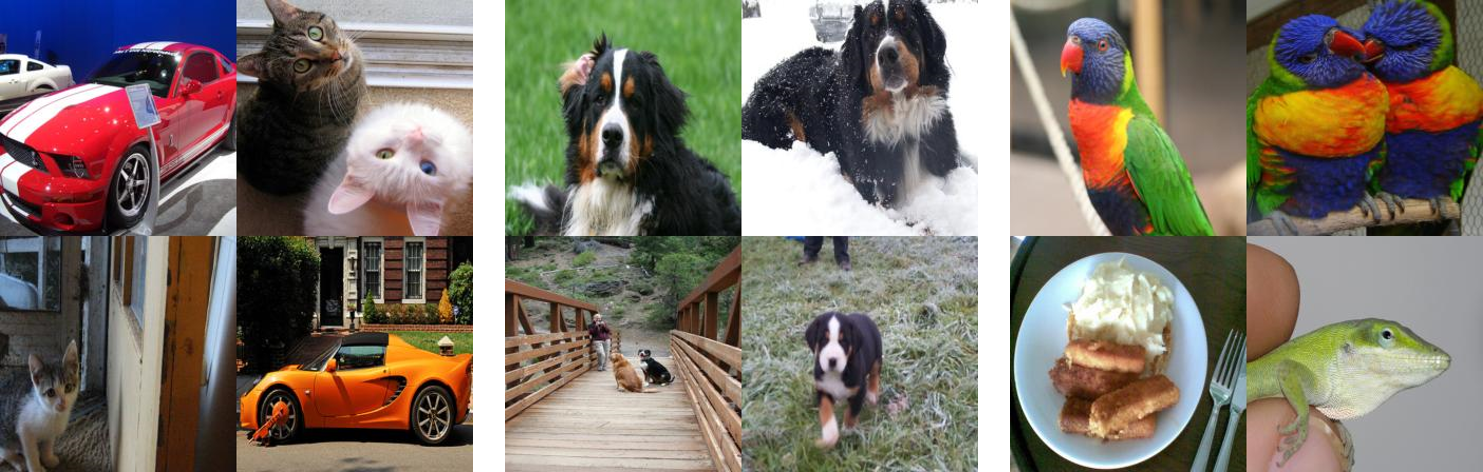}
    \caption{One sample mosaic for each of the regarded datasets (cf. Section~\ref{sec:experiments}). On the left the mosaic comprises the ImageNet classes ``tabby'' and ``sports car'', in the middle ``Bernese Mountain Dog'' and ``Greater Swiss Mountain Dog'', and on the right the classes ``lorikeet'', ``mashed potato'', and ``American chameleon''.}
    \label{fig:mosaic_samples}
    \Description{One example mosaic from each of the regarded datasets. The first mosaics shows a 2 by 2 grid structure with sports car and cat-images, the second one shows the very similar looking Greater Swiss Mountain Dog and Bernese Mountain Dog classes and the third one shows two images of lorikeets, a type of bird, a plate with mashed potatoes and a chameleon. }
\end{figure*}

The proposed approach adapts the procedure from the original Focus paper \cite{arias2022focus}.
The mosaics used to test and evaluate various saliency methods are constructed of four images: two from the assigned target class and two from different classes within the same dataset. All images are selected randomly from their specific classes. The images are arranged in random positions in the $2 \times 2$ grid without overlap. To maintain consistency of visual patterns between mosaics and the training data, the individual images are scaled to a uniform size of $224\times224$ pixels.
Accordingly, the mosaics have a resolution of 448 $\times$ 448 pixels.
Because the individual images of the mosaics are part of the training data, the noise introduced by them is ensured to fall within the distribution of the training data. Figure \ref{fig:mosaic_samples} shows an example mosaic for each dataset considered in this paper.
The datasets used for mosaic construction for the experiments are described in detail in Section~\ref{sec:experiments}.

\begin{table}[btp]
  \caption{\ac{XAI} evaluation metrics proposed in this work. All of them can be calculated with the true positives, true negatives, false positives, and false negatives defined in Section~\ref{ssec:true_false_FI}.}
  \label{tab:def_metrics}
  \begin{tabular}{ccl}
    \toprule
    Evaluation Metrics&Formula\\
    \midrule
    Precision (Focus score) & $\frac{tp}{tp + fp}$ \\
    Sensitivity (Recall) & $\frac{tp}{tp + fn}$ \\
    Specificity & $\frac{tn}{tn + fp}$ \\
    False-Negative-Rate & $\frac{fn}{tp + fn}$ \\
    False-Positive-Rate & $\frac{fp}{tn + fp}$ \\
    Accuracy & $\frac{tp + tn}{tp + tn + fp + fn}$ \\
    F1-Score & $\frac{2 \cdot \text{precision} \cdot \text{sensitivity}}{\text{precision} + \text{sensitivity}}$ \\
  \bottomrule
\end{tabular}
\end{table}

\subsubsection{True and False Feature Importance}\label{ssec:true_false_FI}

For a more holistic evaluation of \ac{XAI} methods, further metrics are defined in addition to the Focus score---the precision for \ac{FI}---by considering negative \ac{FI}. In general, pixels with a positive feature attribution value contribute to the prediction of the target class and pixels with a negative feature attribution value contribute to the prediction of the other classes. Here, both are taken into account for the specification of true positives and negatives as well as false positives and negatives related to the \ac{FI} on the mosaics. However, true and false \ac{FI} can only be approximated, because the pixel-wise ground truth is unknown. This is done as follows:

\begin{itemize}
    \item true positive ($tp$): positive \ac{FI} on the images of the target class
    \begin{displaymath}
        tp = \sum_{x,y}{\operatorname{FI}(c_\mathrm{img}(x,y)=c_\mathrm{target} \land \operatorname{FI}>0)}
    \end{displaymath}
    where $x,y \in \lbrace 0, 1 \rbrace$ are the image coordinates, i.e., $(0,0)$ is the image on the bottom left, $(0,1)$ is bottom right, $(1,0)$ is top left, and $(1,1)$ is top right. \ac{FI} is the feature importance in the entire mosaic, $c_\mathrm{img}$ is the class label at position $(x,y)$, based on the four images used to create the mosaic and $c_\mathrm{target}$ is the target class for the \ac{FI}. The computation of the \ac{FI} depends on the saliency method under investigation. 
    \item false positive ($fp$): positive \ac{FI} on the images that do not belong to the target class
    \begin{displaymath}
        fp = \sum_{x,y}{\operatorname{FI}(c_\mathrm{img}(x,y)=\lnot c_\mathrm{target} \land \operatorname{FI}>0)}
    \end{displaymath}
    \item false negative ($fn$): negative \ac{FI} on the images of the target class
    \begin{displaymath}
        fn = \sum_{x,y}{\operatorname{FI}(c_\mathrm{img}(x,y)=c_\mathrm{target} \land \operatorname{FI}<0)}
    \end{displaymath}
    \item true negative ($tn$): negative \ac{FI} on the images of other classes
    \begin{displaymath}
        tn = \sum_{x,y}{\operatorname{FI}(c_\mathrm{img}(x,y)=\lnot c_\mathrm{target} \land \operatorname{FI}<0)}
    \end{displaymath}
\end{itemize}

Thus, by also considering negative \ac{FI}, true and false negatives can be calculated in addition to true and false positives, so that a full confusion matrix can be defined. This enables the computation of metrics commonly applied in classification tasks, specifically for assessing saliency methods (cf. Table~\ref{tab:def_metrics}). With the additional metrics, \ac{XAI} methods can therefore be evaluated more comprehensively.


Please note that not all \ac{XAI} methods provide negative \ac{FI}. Accordingly, the additional metrics can only be calculated for B-cos, \ac{IntGrad}, \ac{LRP} and \ac{SHAP}. The other \ac{XAI} methods can only be evaluated using the precision metric. An approach to examine the suitability of the metrics for evaluating the \ac{XAI} methods is explained in the next section.

\subsection{Evaluation Approach}
Despite the absence of ground truth for evaluating the explanation methods and, consequently, the saliency metrics, the evaluation of certain properties of such metrics can still be conducted. Given the analogous challenges of lacking ground truth in psychometric approaches, corresponding evaluation procedures can be adapted to saliency maps, as proposed in \cite{tomsett2019sanityformetrics}. Two fundamental concepts are \emph{validity}, i.e., the extent to which a test or variable measures what it is intended to measure,  and \emph{reliability}, i.e, the consistency of results a test produces. While a reliable test does not guarantee validity, reliability is a necessary condition for validity \cite{murphy2004pychological-testing}.
In psychometric testing, the scenario is usually described by raters (e.g., psychologists) administering tests to a patient, where different types of reliability can be evaluated to assess whether a metric produces reliable and thus, possibly valid results.
In this paper, two adapted reliability tests from \cite{tomsett2019sanityformetrics} are considered.

\subsubsection{Inter-rater Reliability}

For the purpose of selecting a metric to choose between various saliency methods, this metric ideally yields the same ranking of saliency methods across all images of a dataset. When the ranking of saliency methods remains consistent across all (or most) test images, it is highly likely that the ranking for new images will be the same as well. This makes it easier to identify the best performing saliency method for future tasks. This paradigm can be compared to \emph{inter-rater reliability}, where the images can be regarded as different raters administering a battery of tests to be scored by the saliency methods \cite{tomsett2019sanityformetrics}. 
Intuitively, each image (rater) produces a ranking of saliency methods via the respective metric.
This ranking can then be checked for agreement over all images (raters) across a dataset.
Krippendorff's $\alpha \in [-1, 1]$ is a common statistic used to assess agreement between raters \cite{krippendorff2004alpha}.
It is calculated as
\begin{math}
    \alpha = 1 - \frac{D_o}{D_e}
\end{math}
, where $D_o$ denotes the disagreement observed and $D_e$ denotes the disagreement by chance.
A value of $\alpha = 1$ signifies perfect agreement in the ranking of saliency methods, $\alpha$ close to $0$ indicates random rankings, and $\alpha < 0$ indicates systematic disagreement.
The implementation used is the one by \citet{castro2017fastKrippendorff}.

\subsubsection{Inter-method Reliability}

The saliency precision can also be used to identify images or classes that are particularly challenging to classify within a given dataset \cite{arias2022focus}.
In order to utilize an evaluation metric for this objective, difficult classes and images should be found consistently.
This consistency should be independent of the saliency method employed, as the model used remains the same across all methods.
Such a desideratum can be compared to \emph{inter-method reliability}, which can be quantified using Spearman's $\rho$ \cite{tomsett2019sanityformetrics}, which measures whether the relation of two variables $X, Y$ can be described via a monotonic function (i.e., an increase in $X$ also results in an increase in $Y$ \cite{Spearman1904}).
Spearman's $\rho$ can be calculated as the Pearson correlation $p$ between the ranks of $X$ and $Y$, resulting in
\begin{math}
    \rho = p_{R(X), R(Y)} = \frac{\text{cov}(R(X), R(Y))}{\sigma_{R(X)} \cdot \sigma_{R(Y)}}
\end{math},
where $\sigma_{R(X)}$ and $\sigma_{R(Y)}$ denote the standard deviations of the rank variables $R(X)$ and $R(Y)$, respectively, and $\text{cov}(R(X), R(Y))$ the covariance of the rank variables \cite{Myers2013researchDesign}.
A high value of $\rho$ indicates that the saliency methods exhibit agreement in their variations across different images. Consequently, images with high (low) saliency metric scores for one method will consistently receive similarly high (low) scores across all saliency methods.

\section{Datasets}\label{sec:experiments}
In the following, general information about the experiment setup is given, before the used mosaic datasets are described in more detail.

To compute the metric scores, the relative magnitude of the \ac{FI} is used.
For visualisation sake, the saliency maps are normalized via max-scaling, mapping them to the interval $[-1,1]$. 
This is the straightforward extension of the normalization used by \citet{arias2022focus} to also work for saliency methods with negative \ac{FI}.
This normalization preserves the 0-point of the saliency maps and leaves the proposed metrics unchanged.

We evaluate the metrics on two different neural network architectures, with the comparatively small VGG architecture with a depth of 11 layers \cite{simonyan2015vgg} and the larger ResNet architecture with a depth of 50 layers \cite{he2015resnet}.
Since the VGG architecture contains batch normalization, its implementation differs slightly between the B-cos-version and the conventional one.
To remove all bias terms in their networks, the authors of B-cos change the batch normalization to not contain a centering operation, resulting in a so-called ``uncentered'' batch normalization \cite{böhle2023bcos}.

In general, it is difficult to disentangle the performance of the model and the performance of the explanation methods. Specifically, incorrectly distributed \ac{FI} and thus, a low saliency metric score, could indicate either a high performing model with a low-fidelity explanation method (with feature attribution distributed evenly across a mosaic) or a bad model and a high-fidelity explanation method. To distinguish between these cases, the saliency scores and the model performance must always be considered in combination.

To test the reliability over different datasets and models and the amount of information provided by the saliency metrics, the scenarios described in the following subsections are tested. The hyperparameters used for the \ac{XAI} methods can be found in Appendix~\ref{a-ssec:hyperparameters}.

\subsection{Corner Cases with Small Datasets}
To establish the overall behavior of the proposed metrics, their performance and coherence with expectation can be tested in simple corner cases, for which clear expectations can be formulated.
As corner cases, for which the metric behavior can be predicted, two datasets are used with two classes and 100 mosaics per class each.
For these datasets, the classes are chosen from the ones represented in ImageNet, with images for the mosaics chosen from all class images at random.
As all models are pretrained on ImageNet and a benchmark of their performance is most informative without any changes, the models are not fine-tuned on the specific datasets. 
This evaluation approach limits options regarding datasets, as only ones built from ImageNet (or at least with the same classes) can be used.
Otherwise, fine-tuning of models or relabeling of classes would be necessary. This is contrary to \citet{arias2022focus}, who adapt the network architecture to the
number of classes in the dataset under consideration and thus, need to fine-tune their models.
Our approach provides an unbiased assessment of popular models but also complicates the reporting of accuracy, as models trained on ImageNet might have learned high-level features like the difference between cats and dogs but not the specific difference between certain dog breeds, resulting in a low top-1 accuracy but in a high top-k accuracy for k $>$ 1.
Thus, both top-1 and top-5 accuracy need to be considered to make sure that the used models have learned relevant features to classify the datasets correctly.

A further difference to \cite{arias2022focus} is the creation of the mosaics:
For the corner cases in Sections \ref{sssec:easy-dataset} and \ref{sssec:difficult-dataset}, samples from the training set of the ImageNet subset~\cite{imagenet-object-localization-challenge} are used for the mosaic construction, instead of only test data samples (where `training' and `test' refer to the corresponding partitions of this dataset).
Since the aim of this work is to test the proposed metrics for the evaluation of different \ac{XAI} methods (in contrast to e.g. performance evaluation of the networks), no relevant effects of leakage are expected.
This assumption was confirmed in experiments with unseen datapoints of the same dataset.

\subsubsection{Easy to Distinguish Classes} \label{sssec:easy-dataset}
In the first corner case, the mosaics for the saliency metrics are created with two ImageNet-classes, which are expected to consist of very dissimilar features and thus, should be easily distinguishable by the pre-trained models.
The target classes ``tabby'' and ``sports car'' are used and the dataset is referred to as the Cars/Cats dataset in the following.
Since there should be (nearly) no overlap between the relevant features for both classes, next to perfect saliency metric scores are to be expected.
One sample mosaic for this dataset can be seen in Figure~\ref{fig:mosaic_samples} on the left.


\subsubsection{Difficult to Distinguish Classes} \label{sssec:difficult-dataset}
The second dataset consists of mosaics built from two classes that look similar to laypeople and have strongly overlapping features.
The classes chosen for this dataset are ``Greater Swiss Mountain Dog'' and ``Bernese Mountain Dog''. The dataset is referred to as the Mountain Dogs dataset. 
For these mosaics, it is expected that the models will not be able to separate these classes, resulting in near-random performance and saliency metrics.
One sample mosaic for this dataset can be seen in Figure~\ref{fig:mosaic_samples} in the middle.

\subsection{ImageNet}\label{ssec:imagenet}
After testing the saliency metric behavior for corner cases, the third dataset uses all classes of the ImageNet dataset \cite{imagenet} as described in \cite{arias2022focus}. The mosaics constructed by \citet{arias2022focus} are available online and were used here.
Compared to the other datasets, this dataset better represents most real-world computer vision tasks.
For all of the 1,000 ImageNet-classes, mosaics are created, but due to hardware and runtime constraints, only ten mosaics per target class are feasible, resulting in 10,000 mosaics overall.
The hardware used for the experiments and resulting runtime for this dataset is described in Appendix~\ref{a-sec:runtime}.
For the ImageNet dataset, the mosaics once again contain two target class images, with the other two images being chosen at random from all other possible classes, as can be seen on the right side of Figure~\ref{fig:mosaic_samples}.

\section{Results} \label{sec:results}
\begin{figure*}[t]
    \includegraphics[width=\textwidth]{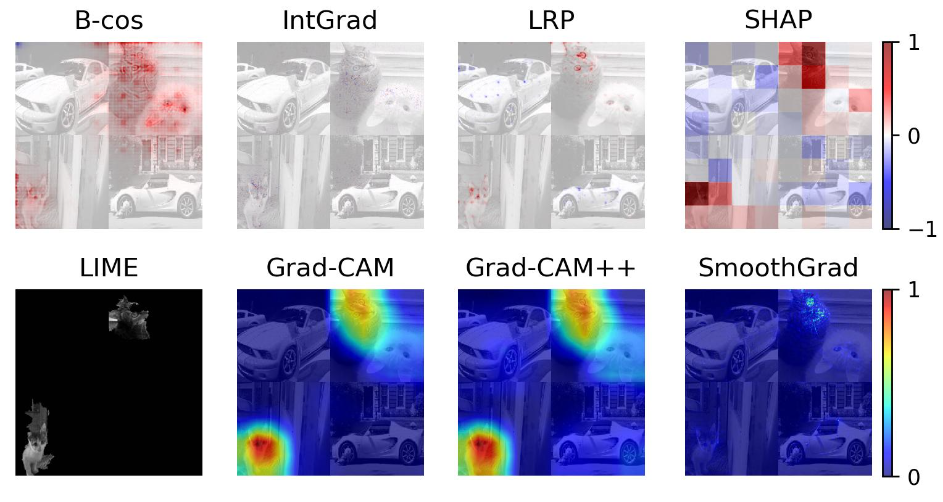}
    \caption{One sample heatmap by each saliency method for the first mosaic shown in Figure \ref{fig:mosaic_samples} of the Cars/Cats dataset. The explanations are created for ResNet50 for the target class ``tabby''. The upper row shows heatmaps for methods providing positive and negative \ac{FI}, the lower one for methods with only positive \ac{FI}. \ac{LIME} uses a binary mask to highlight relevant image pieces, thus a binary masking of the original image is shown here. Similar results for VGG11 are presented in Figure~\ref{fig:mosaic_results_vgg11} in the appendix.}
    \label{fig:mosaic_results}
    \Description{This figure shows heatmap results for the saliency methods used in this paper for a Cars/Cats mosaic. One can see that these results differ in the area they mark as relevant between the methods, although most methods mainly highlight parts of the cats as important for the prediction of the class "tabby". B-cos additionally highlights the upper left corner of the image.}
\end{figure*}

In this chapter, the results and findings of the proposed saliency metrics are discussed. At first, the results for inter-rater and inter-method reliability are summarized, followed by some general findings with the saliency methods and metrics in Section~\ref{ssec:general-findings}.
To give a better intuition for these results, Figure~\ref{fig:mosaic_results} provides an example of how the saliency maps differ between the saliency methods for ResNet50. The same can be seen for VGG11 in Figure~\ref{fig:mosaic_results_vgg11} in the appendix.
A more detailed view and discussion of the results can be found in the Appendix~\ref{a-sec:experiments}. 

\subsection{Inter-rater Reliability} \label{sec:inter-rater-reliability}
Krippendorff's $\alpha$ can be used as a metric for inter-rater reliability, indicating whether the ranking of \ac{XAI} methods by a saliency metric is stable over all (or most) of the images in a dataset.
Detailed results can be found in appendix \ref{a-ssec:inter-rater-reliability}.
In these, some tendencies emerge: The consistency of the saliency method ranking depends on the model type.
In the experiments, ResNet50 almost always receives higher $\alpha$-values than the VGG11-model.

SmoothGrad, \ac{LIME}, \ac{Grad-CAM}, and \ac{Grad-CAM}++ provide only positive \ac{FI}, thus only the precision-reliability (the reliability of the original Focus metric \cite{arias2022focus}) can be evaluated for all used saliency methods.
For the datasets, the easier the models can distinguish between classes, the more reliable the precision ranking becomes, with the highest values for ResNet50 for the Cars/Cats dataset with $\alpha = 0.88$ and for ImageNet with $\alpha = 0.71$.
The highest $\alpha$-values for VGG11 are below $0.6$, thus underpinning that the saliency method performance (and metric reliability) depends on the model type.

When the precision-reliability is evaluated for only B-cos, \ac{LRP}, \ac{IntGrad} and \ac{SHAP}, the results are similar, but for these methods, additional saliency metrics can be calculated with negative \ac{FI}.
They follow similar trends: The easier the dataset, the higher the reliability, and in general higher reliability for ResNet50 than for VGG11, except for the false-positive-rate and specificity on ImageNet.
Although not perfect, the inter-rater reliability for all classes of ImageNet (with values between $0.49$ and $0.85$) shows that the saliency metrics produce consistent results, thus enabling a user to choose between different saliency methods.
Overall, the reliability of sensitivity, false-negative-rate, false-positive-rate, specificity, accuracy, and F1-score is higher than for precision, showing the added benefits of these metrics.

\subsection{Inter-method Reliability} \label{sec:inter-method-reliability}
Spearman's $\rho$ correlation between the results of the metrics for different saliency methods can be used to examine whether mosaics are consistently difficult to explain correctly for all methods.
For $\rho$, some dependencies emerge: The correlation values for one metric differ between models on the same dataset, between different datasets for the same model and between the different metrics.
For more detailed results, see Section~\ref{a-ssec:results} and Figure~\ref{fig:corrs} in the appendix.
While most correlation values are rather low ($< 0.8$), in some cases for certain methods, correlations close to $1$ can be seen, especially for the datasets for which the used classes were expected to be difficult to distinguish (especially on the Mountain Dogs dataset for \ac{Grad-CAM}, \ac{Grad-CAM}++ and \ac{SHAP}).
On these datasets, all \ac{XAI} methods do not perform well based on the saliency metrics, thus possibly indicating a joint failure of certain saliency methods.
For the other datasets, no clear correlation pattern emerges, with correlations $< 0.8$.
Overall, the performances of the \ac{XAI} methods can be highly correlated between some of them, given that the model is not able to distinguish well between different classes, while for more diverse datasets, the performances of the \ac{XAI} methods are not strongly correlated.
This could be paraphrased as: ``\emph{The saliency methods tend to work individually but some of them fail jointly.}''


\subsection{General Findings} \label{ssec:general-findings}
Additional to the more specific findings above, some general tendencies for the saliency methods can be identified.

B-cos highlights the upper left corner of images, possibly because the bias was removed in the network architecture, forcing the network to ``create its own bias'' via mostly irrelevant but stable features like image edges \cite{böhle2023bcos} (see Figure \ref{fig:mosaic_results} for an example).

\begin{figure*}[tbp]
  \centering
  \begin{subfigure}[t]{0.45\linewidth}
    \includegraphics[width=\linewidth]{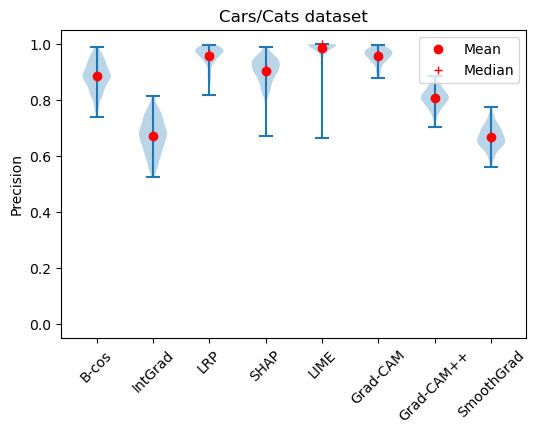}
    \caption{Precision for all saliency methods on the Cars/Cats dataset.}\label{sfig:precision-tabby}
  \end{subfigure}
  \hfill
  \begin{subfigure}[t]{0.45\linewidth}
    \includegraphics[width=\linewidth]{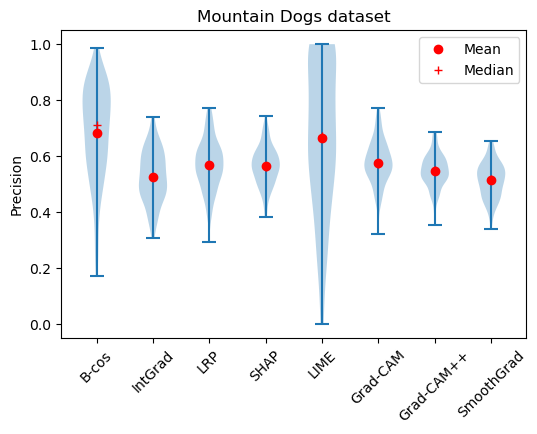}
    \caption{Precision for all saliency methods on the Mountain Dogs dataset.}\label{sfig:precision-dog}
  \end{subfigure}
  
  
  \begin{subfigure}[t]{0.45\linewidth}
    \includegraphics[width=\linewidth]{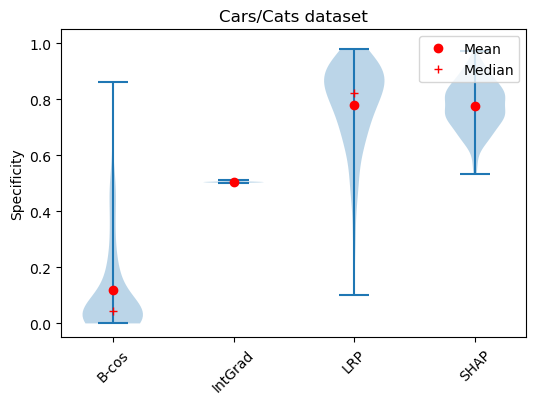}
    \caption{Specificity for saliency methods with negative \ac{FI} on the Cars/Cats dataset.}\label{sfig:specificity-tabby}
  \end{subfigure}
  \hfill
  \begin{subfigure}[t]{0.45\linewidth}
    \includegraphics[width=\linewidth]{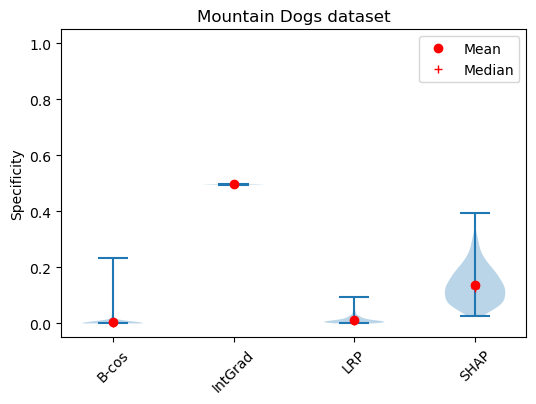}
    \caption{Specificity for saliency methods with negative \ac{FI} on the Mountain Dogs dataset.}\label{sfig:specificity-dog}
  \end{subfigure}
  \caption{Exemplary results for precision and specificity for ResNet50 on the datasets with easier and more difficult to distinguish classes. Higher values are better. Note that specificity can only be calculated for methods which provide negative \ac{FI}.}\label{fig:results}
  \Description{This figure shows violin plots for the saliency methods for their precision and specificity for the Cars/Cats dataset, the precision is higher with a lower variance than for the Mountain Dogs dataset. For specificity, all methods perform bad with lower variance for the second one of these datasets, while for Cars/Cats, only B-cos performs bad, while the \ac{IntGrad} specificity is the same for both datasets.}
\end{figure*}

\ac{IntGrad} does seem to yield mostly random performances in the metrics. Together with a good balance between positive and negative \ac{FI}, this results in metrics close to $0.5$ (see Figures~\ref{sfig:specificity-tabby} and \ref{sfig:specificity-dog}).
As can be seen in Figure~\ref{sfig:precision-tabby}, the precision for the Cars/Cats dataset for \ac{IntGrad} is above $0.5$, showing a performance better than random guessing as indicated by the other metrics.
Visually, \ac{IntGrad} explanations do mainly look like noise that seems to be stronger on the relevant image parts.

While some methods in theory provide negative \ac{FI}, the magnitude of their positive importance is higher than the negative one, thus yielding misleading interpretations for some of the metrics when inspected on their own.
This is illustrated, for example, by the precision and specificity in Figure~\ref{fig:results}: B-cos provides a high precision, but a low specificity, because it barely provides any negative \ac{FI}-values and is tailored towards the correct attribution of positive \ac{FI}.
This imbalance towards positive \ac{FI} is especially prevalent when the classes in the mosaics are more difficult to distinguish (see Figures~\ref{sfig:specificity-tabby} and \ref{sfig:specificity-dog}).

Underpinning the initial intuition in creating the datasets in Section~\ref{sec:experiments}, the precision is significantly lower for Mountain Dogs than for Cars/Cats (Figure~\ref{fig:results}), with the precision for ImageNet somewhere in between.

None of the tested methods provide good results in all of the metrics over different datasets, despite a sufficiently high classification accuracy for all datasets, showing that the models have learned relevant features (see Appendix~\ref{a-sec:accuracy}).
While B-cos for example fared well in mean and median-performance for precision for all datasets, its specificity consistently produced values close to $0$ (due to the higher prevalence of positive \ac{FI}, see above and Figure~\ref{sfig:precision-tabby} compared to Figure~\ref{sfig:specificity-tabby}).
Here, it is important to note that the inter-rater reliability only measures the agreement over the saliency method ranking within a given metric but does not indicate that the different metrics lead to the same ranking of saliency methods.
Complementary to the ``eye-check'' for B-cos and other methods as above, this aspect could be explored via Krippendorff's $\alpha$ between different metrics on the same dataset and model. 
Since the ``eye-check'' of the saliency method ranking between the metrics already showed that metrics usually produce different rankings and no saliency method performs well in all of them, this aspect was not explored further (Appendix~\ref{a-ssec:imagenet-metrics}).
An additional aspect of reliability-evaluation could be the agreement over mean and median performances of the saliency methods over different datasets, as this would show whether some methods consistently produce better performances on different datasets.
As an analysis of this type of reliability just underpinned the previous results of ResNet50 providing more reliable results and method rankings mostly differing between datasets, a detailed discussion is ommitted here.
For all of the datasets, large variances in the metrics can be discerned, thus indicating that some images produced almost perfect scores while others received scores towards the other end of the scale.

In classical literature, it has been long known that a single metric is not sufficient and multiple metrics are necessary to obtain a reliable assessment of a method, especially when some sort of unbalanced dataset is used \cite{Tharwat2021classificationMetrics}.
In this paper, the existence of such an imbalance in the saliency methods was shown.

For ImageNet, the explanation methods recommended---at least for the properties of correctness and contrastivity---are \ac{LRP} and \ac{SHAP}, although both show clear weaknesses (see appendix \ref{a-ssec:imagenet-metrics}).
\ac{LRP} performs slightly better in some cases, but overall the performance of \ac{SHAP} is more consistent for ResNet50 compared to \ac{LRP}, especially for specificity and false-positive-rate.
For VGG11, the variance of \ac{LRP} is lower than for ResNet50, thus rendering \ac{LRP} the best explanation method for this model for the ImageNet-mosaics.
While a recommendation for saliency methods for a specific model and use-case can be made with the proposed metrics, the metric values and variances also show that no method performs to complete satisfaction (as, for instance, the highest mean-specificity for ResNet50 on the ImageNet-dataset is below $0.6$), prompting more research for improved saliency methods.

Overall, the \ac{XAI} methods perform differently in different scenarios, possibly because their underlying concepts of what constitutes important features differ \cite{watson2020conceptualChallenges}.
The reliability assessment of the saliency metrics shows that the proposed metrics can help to choose between saliency methods, although it should be noted that a single saliency metric does not yield sufficient information for this choice for a given use-case.
Instead, the combination of dataset, model, and \ac{XAI} method needs to be evaluated to receive a meaningful assessment of the properties of saliency methods.

\section{Discussion} \label{sec:discussion}

This paper examined the question whether the relevant features (i.e., the positive feature importance) for a class are actually located on the images of this class and if this fact can be utilized to define sensible metrics for evaluating saliency methods. Overall, this assumption holds, however, the introduced metrics are not exactly intuitive. They range between $0$ and $1$, where $1$ can usually not be reached and $0.5$ corresponds to random guessing.
Additionally, if images of classes with very similar features are present in the mosaic (cf. Mountain Dogs dataset), the assumption is likely to be violated.

Another challenge for the metrics arises when, for example, all images of one class have the same background and this background is only present in this class in the dataset under consideration. In such a case, the metrics provide high (resp. low) scores for the generated explanations, but the explanations would show some sort of bias within the model when inspected by humans. It is important to note that the saliency metrics do not contain information about the visual quality of explanations and rather correspond to a sanity check. For the saliency methods examined, higher (resp. lower) scores are always considered better, nevertheless, the metrics could be outsmarted: for instance, an explanation method that only attributes relevance to a single image pixel would lead to perfect scores but does not provide helpful information at all.

There are also some limitations of the methodology that need to be addressed: the random choice of images used for creating mosaics may introduce bias, e.g., by selecting images that are too (dis)similar and especially easy or difficult to distinguish. To mitigate this, the experiments, including mosaic creation, can be repeated multiple times or the number of generated mosaics can be increased. However, with the given number of mosaics, such effects are expected to balance out within the datasets used in this paper without exceeding runtime limits. Additionally, it is worth emphasizing that the B-cos method uses different model weights and activation functions than the other \ac{XAI} methods, which could raise concerns about the direct comparability of explanation results.

Viewed from the outside, there is also the meta-level problem: saliency methods are used to understand and evaluate black box \ac{ML} models. Saliency metrics are then used to evaluate the saliency methods and these metrics are then checked for reliability, etc. From a practical viewpoint, low-level information about which \ac{XAI} methods to choose needs to be available without excessive amounts of work for evaluating different \ac{XAI} approaches. But since there is no ground truth that can be used to verify statements at any level, the entire framework remains shaky. On the other hand, as there can probably not be a full ground truth explanation of a black-box-model that is different from just the model itself, it is necessary to employ the methods at hand to illuminate the underlying complexities at least to some extent. Therefore, it is crucial to always explicitly state the main assumptions of \ac{XAI} methods and possible bias that may occur when using them, as these aspects are fundamental for selecting a suitable method for the respective model and dataset.

For future research, it would be interesting to extend the list of metrics to address further \ac{XAI} properties listed in \cite{Nauta2023XAIevalsurvey}. In addition, the proposed metrics could be applied to \ac{XAI} methods on specific benchmark datasets to analyze and evaluate the resulting explanations.

\section{Summary} \label{sec:summary}
In this paper, new objective evaluation metrics for saliency methods were developed based on the definition of true (false) positive and negative \ac{FI} in image mosaics.
This definition required the assumption that evidence towards a certain class would be more prevalent in images of this class than in others, enabling the saliency metrics to use image mosaics as the basis of their calculation, mimicking common classification evaluation metrics.
To test these metrics, datasets with mosaic images were created, small ones to evaluate corner cases with especially easy or difficult to distinguish classes and a larger one based on all classes of ImageNet.

For the practical use of a measurement, its validity---with its necessary condition of reliability---is crucial.
Via inter-rater and inter-method reliability, the proposed metrics were established to be reliable in most cases, with the overall results showing that the performance of common saliency methods depends on the \ac{ML} model and used dataset.
As no clear correlation between the different saliency method results could be found, it seems that the saliency method performance also depends on the specific image being explained and goes beyond just single images being easy or difficult to explain.
Due to their high inter-rater reliability, the proposed saliency metrics can be used to choose between different saliency methods for a specific use-case, although, due to the method's focus on positive \ac{FI}, more than a single metric needs to be taken into account where possible.
As these methods only assess the contrastivity and correctness of saliency metrics, we look forward to proposals of objective, reliable, and valid evaluation metrics for other properties of \ac{XAI} methods and further reliability evaluations of other saliency metrics.

\begin{acks}
Parts of this paper were created with the help of a company-specific implementation of ChatGPT (3.5 turbo).
It was used to create LaTeX-code for tables and figures and to refine the drafts of some sections.

We thank the reviewers of the FAccT 2024 conference, who helped to improve this paper with their valuable feedback.
This paper is funded in parts by the German Federal Ministry for Economic Affairs and Climate Action under grant no. 19A21040B (project "veoPipe") and by the Fraunhofer Gesellschaft under grant no. PREPARE 40-02702 (project "ML4Safety").
\end{acks}

\bibliographystyle{ACM-Reference-Format}
\bibliography{bibliography}
\newpage
\appendix

\section{Model Analysis}
\subsection{Runtime}\label{a-sec:runtime}

\begin{figure*}[bp]
    \includegraphics[width=0.9\textwidth]{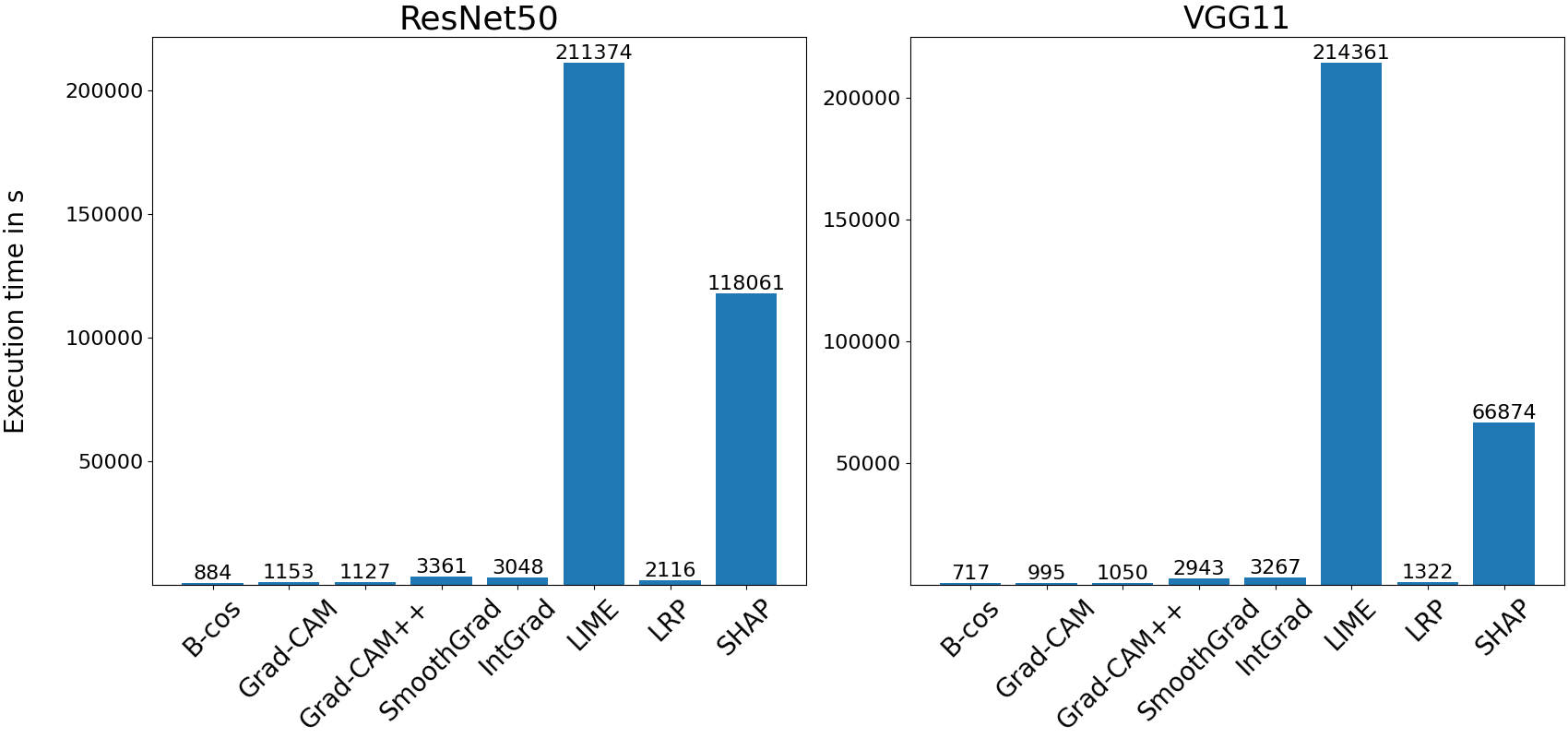}
    \caption{Visualization of the execution time of the different saliency methods in seconds. The time required to generate the saliency maps of every mosaic in the ImageNet dataset (cf. Subsection~\ref{ssec:imagenet}) was measured.}
    \label{fig:runtime}
    \Description{Bar chart visualizing the execution times of the different saliency methods. While the execution time for most methods is below 4000s, the sample-based methods are far slower, with SHAP at 118061s and LIME at 211374s for ResNet50, with similar results for VGG11.}
\end{figure*}

A DGX A100 system with 40 GB of RAM-Memory was used to carry out the experiments described in Section~\ref{sec:experiments}. The code for generating saliency maps with varying saliency methods was executed on a 20 GB MIG slice of an NVIDIA A100 40 GB GPU. The runtimes for generating heatmaps for the ImageNet dataset from Subsection~\ref{ssec:imagenet} with respect to different saliency methods can be seen in Figure~\ref{fig:runtime}. These runtimes highlight the performance benefits of gradient-based methods compared to the sample-based methods of \ac{LIME} and \ac{SHAP}.

\subsection{Accuracy} \label{a-sec:accuracy}
For the model accuracy for mosaic datasets, see Table~\ref{tab:acc-resnet} for ResNet50 and Table~\ref{tab:acc-vgg} for VGG11.
The accuracy is calculated for each class separately, with top-$1$ accuracy denoting whether class 1 or class 2 of a mosaic is predicted as the most likely class, top-$5$ accuracy denoting whether class 1 or class 2 are predicted in the five most likely classes.
As the performance of these models is measured on the mosaic datasets, no true negatives nor false positives are to be expected, as all mosaics do contain images of the relevant classes.
Because of this, only the accuracy can be calculated as a meaningful performance measure.
These results show that both models predict the relevant classes for the mosaics often enough to expect them to have learned the relevant features for these classes.
This aids in the assessment whether the assumption is fulfilled that models should highlight the images in a mosaic that correspond to the target of an explanation.
The accuracy for the Mountain Dogs dataset is higher than for the others, possibly because all of the images in these mosaics contribute to the same target classes.
This explanation is underpinned by the imbalance between positive and negative \ac{FI}, as described in Section~\ref{ssec:general-findings}. 

\begin{table*}[htbp]
    \centering
    \caption{Top-1 and  Top-5 Accuracy for ResNet50 on the mosaic datasets, for the standard and for the B-cos model. Note that only the ImageNet-mosaics contain more than two classes.}
    \label{tab:acc-resnet}
    \begin{tabular}{p{4cm}m{1.25cm}m{1.25cm}m{1.25cm}m{1.25cm}m{1.25cm}m{1.25cm}}
        \toprule
        ResNet50 & top-1, class 1 & top-5, class 1 & top-1, class 2 & top-5, class 2 & top-1, class 3 & top-5, class 3 \\
        \midrule
        Cars/Cats dataset & 0.27 & 0.73 & 0.175 & 0.765 & - & - \\
        Cars/Cats dataset, B-cos & 0.435 & 0.875 & 0.385 & 0.845 & - & - \\
        Mountain Dogs dataset & 0.42 & 0.995 & 0.49 & 0.995 & - & - \\
        Mountain Dogs dataset,  \mbox{B-cos} & 0.345 & 1.0 & 0.41 & 0.995 & - & - \\
        ImageNet & 0.6256 & 0.8683 & 0.0423 & 0.1971 & 0.0447 & 0.2148 \\
        ImageNet, B-cos & 0.476 & 0.8095 & 0.1548 & 0.4727 & 0.1701 & 0.496 \\
        \bottomrule
    \end{tabular}
\end{table*}

\begin{table*}[htbp]
    \centering
    \caption{Top-1 and  Top-5 Accuracy for VGG11 on the datasets for each of the classes represented in the mosaics, for the standard and for the B-cos model. Note that only the ImageNet-mosaics contain more than two classes.}
    \label{tab:acc-vgg}
    \begin{tabular}{p{4cm}m{1.25cm}m{1.25cm}m{1.25cm}m{1.25cm}m{1.25cm}m{1.25cm}}
        \toprule
        VGG11 & top-1, class 1 & top-5, class 1 & top-1, class 2 & top-5, class 2 & top-1, class 3 & top-5, class 3 \\
        \midrule
        Cars/Cats dataset & 0.32 & 0.52 & 0.28 & 0.535 & - & - \\
        Cars/Cats dataset, B-cos & 0.375 & 0.805 & 0.435 & 0.86 & - & - \\
        Mountain Dogs dataset & 0.405 & 0.995 & 0.42 & 0.985 & - & - \\
        Mountain Dogs dataset, \mbox{B-cos} & 0.31 & 0.99 & 0.325 & 0.98 & - & - \\
        ImageNet & 0.3451 & 0.5933 & 0.0468 & 0.133 & 0.0447 & 0.1384 \\
        ImageNet, B-cos & 0.3996 & 0.685 & 0.1265 & 0.328 & 0.1386 & 0.3408 \\
        \bottomrule
    \end{tabular}
\end{table*}


\section{Experiments}\label{a-sec:experiments}
\subsection{Hyperparameters}\label{a-ssec:hyperparameters}

Table \ref{tab:hyperparameters} lists the hyperparameters used when executing the \ac{XAI} methods in the experiments. This is done to ensure reproducibility of the results. Hyperparameters that equal the default value and methods that were used only with default hyperparameters are not included in the listing.



\begin{table*}[tbp]
  \caption{List of hyperparameters used when executing the \ac{XAI}-methods during the experiments.}
  \label{tab:hyperparameters}
  \begin{tabular}{ccl}
    \toprule
    Saliency Method&Hyperparameters\\
    \midrule
    \ac{LIME} & $num\_samples = 1000$ \\
    \ac{SHAP} & $num\_samples = 1500$, $super\_pixel\_size = 56$ (equals $8 \times 8 = 64$ superpixels per mosaic) \\
  \bottomrule
\end{tabular}
\end{table*}

\subsection{Results}\label{a-ssec:results}
\begin{figure*}[htbp]
  \centering
  \begin{subfigure}[c]{\textwidth}
    \centering
    \includegraphics[width=0.88\linewidth]{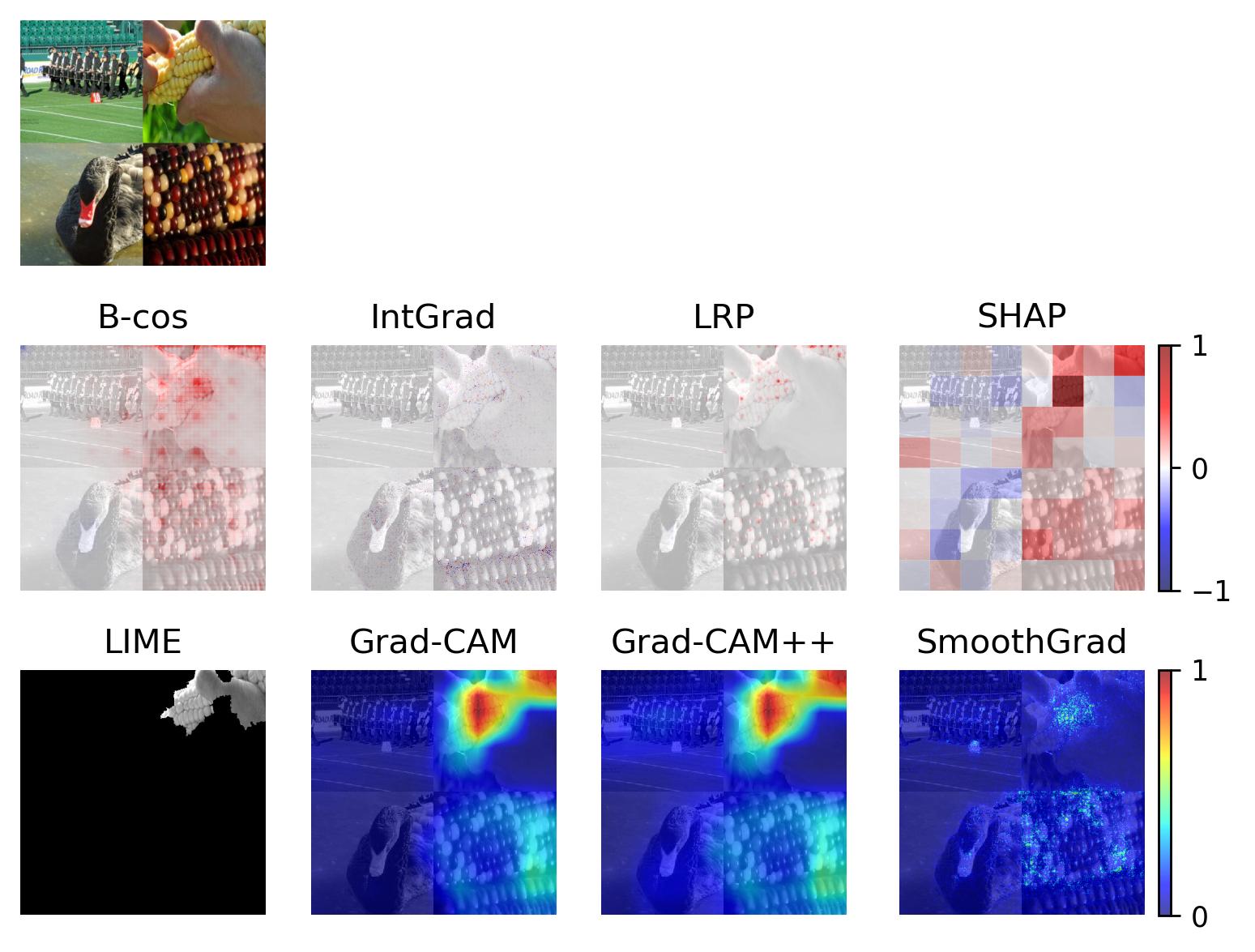}
    \caption{Saliency maps for ResNet50 on an ImageNet mosaic.} \label{sfig:imagenet-resnet-summary}
  \end{subfigure}

  
  \begin{subfigure}[c]{\textwidth}
    \centering
    \includegraphics[width=0.88\linewidth]{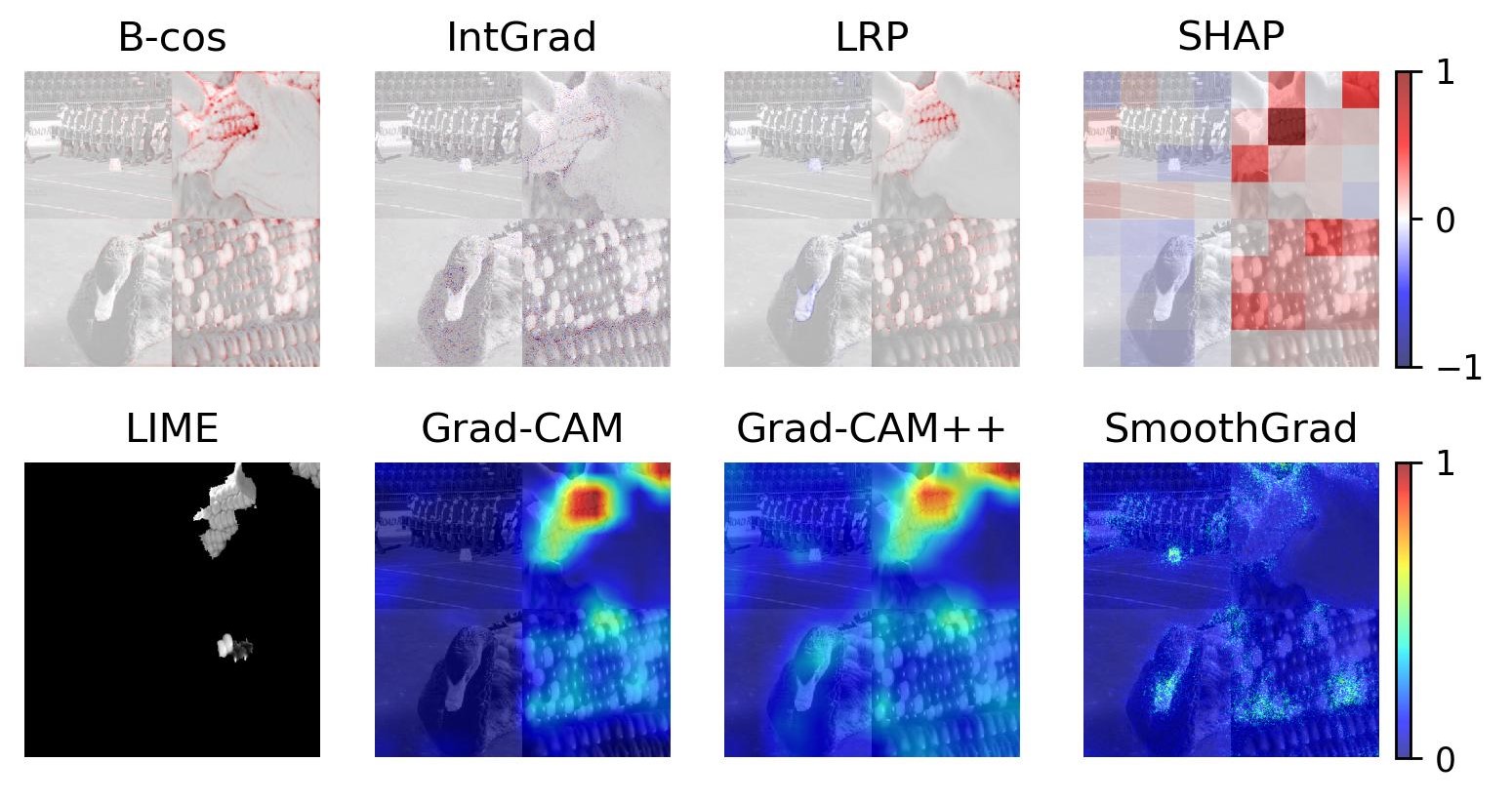}
    \caption{Saliency maps for VGG11 on an ImageNet mosaic.}
    \label{sfig:imagenet-vgg-summary}
  \end{subfigure}
  \caption{Saliency maps for the ImageNet mosaic shown as the first image in Figure \ref{sfig:imagenet-resnet-summary}. The saliency was calculated regarding the target class ``ear, spike, capitulum'', to which the right two images in the mosaic belong.}\label{fig:imagenet-summaries}
  \Description{This figure shows heatmap results for the saliency methods used in this paper for an ImageNet mosaic. One can see that these results differ in the area they mark as relevant between the methods, although these differences appear to be less prominent than between the Cars/Cats mosaics. Additionally, the saliency results differ between the two model types shown here, with VGG11 highlighting smaller areas than ResNet50.}
\end{figure*}

\begin{figure*}[htbp]
  \centering
  \begin{subfigure}[c]{\textwidth}
    \centering
    \includegraphics[width=0.88\textwidth]{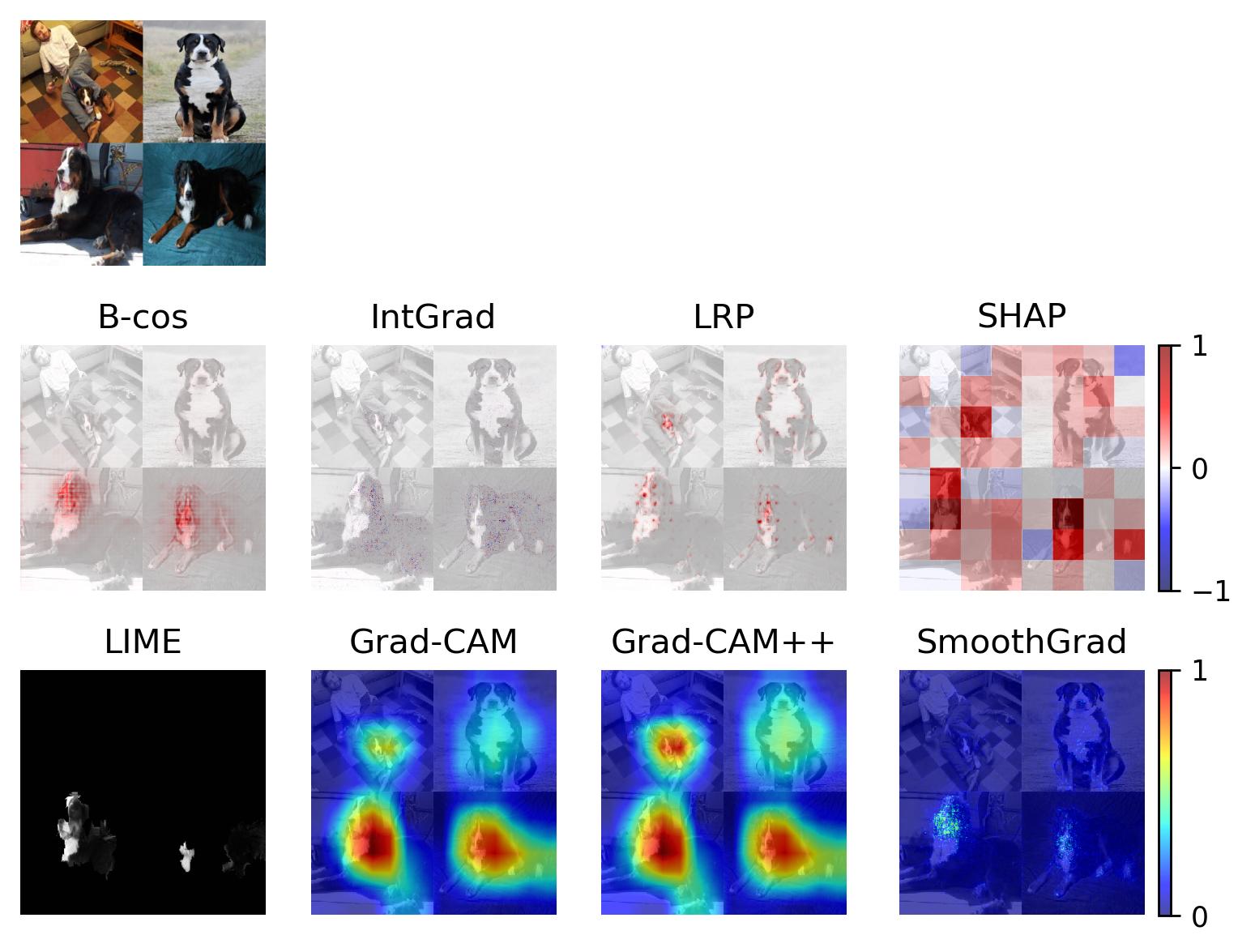}
    \caption{Saliency maps for ResNet50 on a mosaic of the Mountain Dogs dataset.} 
    \label{sfig:dogs-resnet-summary}
  \end{subfigure}
  
  
  \begin{subfigure}[c]{\textwidth}
    \centering
    \includegraphics[width=0.88\textwidth]{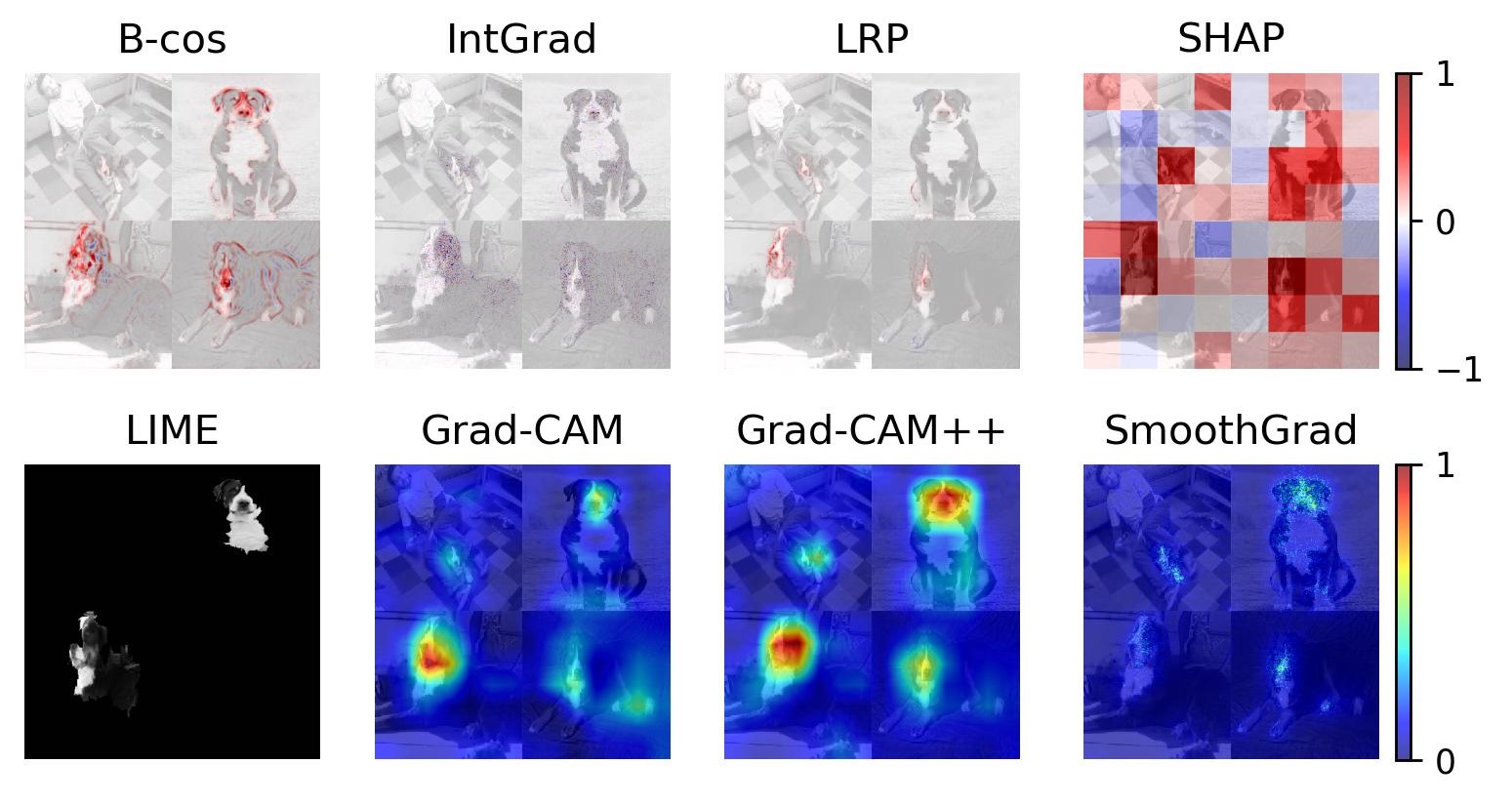}
    \caption{Saliency maps for VGG11 on a mosaic of the Mountain Dogs dataset.}
    \label{sfig:dogs-vgg-summary}
  \end{subfigure}
  
  \caption{Saliency maps for the mosaic shown as the first image in Figure \ref{sfig:dogs-resnet-summary}. The saliency was calculated regarding the target class ``Bernese Mountain Dog'', to which the lower two images in the mosaic belong.}\label{fig:dogs-summaries}
  \Description{This figure shows heatmap results for the saliency methods used in this paper for a Mountain Dogs mosaic. One can see that these results differ in the area they mark as relevant between the methods, although these differences appear to be less prominent than between the Cars/Cats mosaics. Additionally, the saliency results differ between the two model types shown here. Compared to the previously shown results, the importance is more evenly distributed across the mosaic, as these classes were expected to be difficult to distinguish.}
\end{figure*}

For additional saliency maps for ImageNet mosaics see Figure~\ref{fig:imagenet-summaries} and for difficult to distinguish classes see Figure~\ref{fig:dogs-summaries}.
Note that in the second case, the saliency is more evenly distributed across the different images in the mosaics.
For both datasets, differences between the explanations for the model types can be seen.
Compared to Figure~\ref{fig:mosaic_results}, the distinction between classes is less clear in the saliency maps for ImageNet and even less for the Mountain Dogs dataset, resulting in worse saliency metric performance.

\begin{figure*}
    \includegraphics[width=\textwidth]{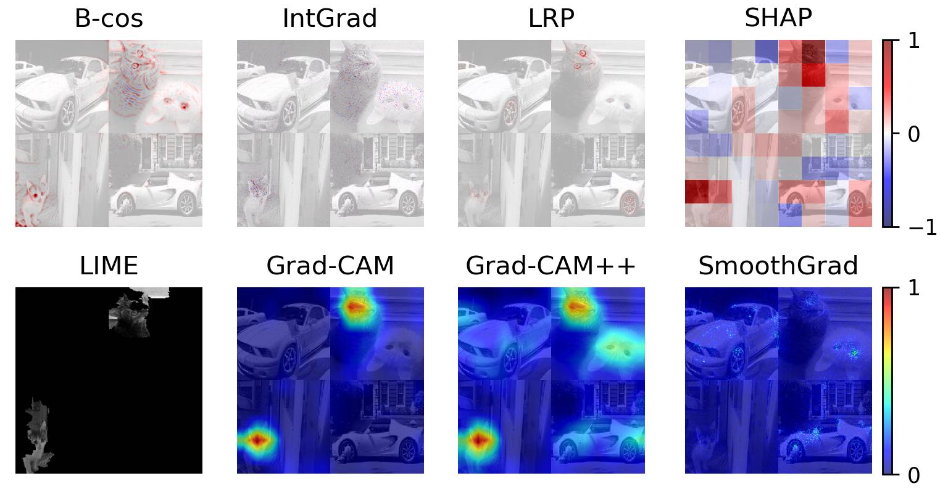}
    \caption{One sample heatmap by each saliency method for the first mosaic shown in Figure \ref{fig:mosaic_samples} of the Cars/Cats dataset, here for VGG11. The upper row shows heatmaps for methods with positive and negative \ac{FI}, the lower one for methods with only positive \ac{FI}. \ac{LIME} uses a binary mask to highlight relevant image pieces, thus a binary masking of the original image is shown here. Note the differences to the explanations for the same image for ResNet50 in Figure~\ref{fig:mosaic_results}.}
    \label{fig:mosaic_results_vgg11}
    \Description{This figure shows heatmap results for the saliency methods used in this paper for a Cars/Cats mosaic, this time for VGG11. One can see that these results differ in the area they mark as relevant between the methods, although most methods mainly highlight parts of the cats as important for the prediction of the class "tabby". Compared to Figure 2 for ResNet50, smaller areas are marked as important for the classification.}
\end{figure*}

\subsection{Metrics for ImageNet} \label{a-ssec:imagenet-metrics}
\begin{figure*}[htbp]
  \centering  
  \begin{subfigure}[t]{0.45\linewidth}
    \includegraphics[width=\linewidth]{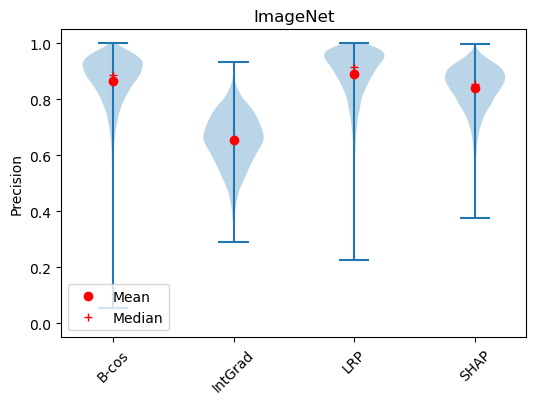}
    \caption{Precision}
  \end{subfigure}
  \hfill
  \begin{subfigure}[t]{0.45\linewidth}
    \includegraphics[width=\linewidth]{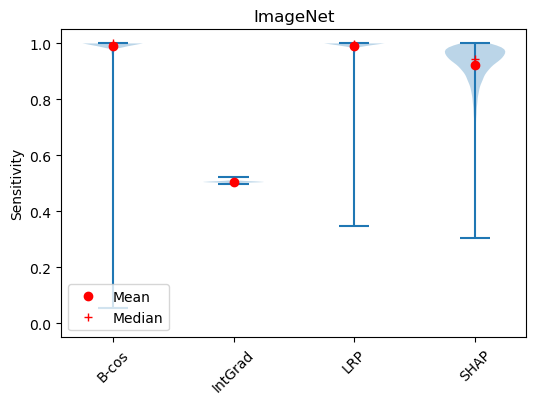}
    \caption{Sensitivity}
  \end{subfigure}

  \begin{subfigure}[t]{0.45\linewidth}
    \includegraphics[width=\linewidth]{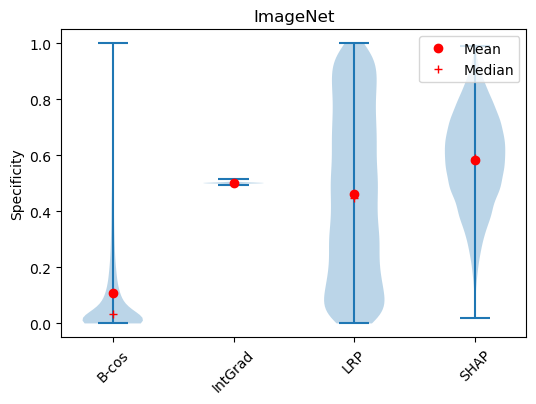}
    \caption{Specificity}
  \end{subfigure}
  \hfill
  \begin{subfigure}[t]{0.45\linewidth}
    \includegraphics[width=\linewidth]{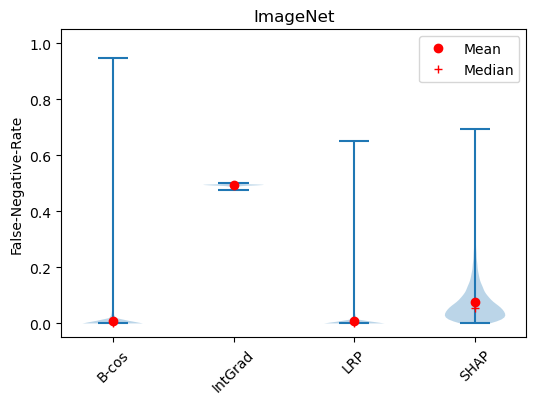}
    \caption{False-Negative-Rate}
  \end{subfigure}

  \begin{subfigure}[t]{0.45\linewidth}
    \includegraphics[width=\linewidth]{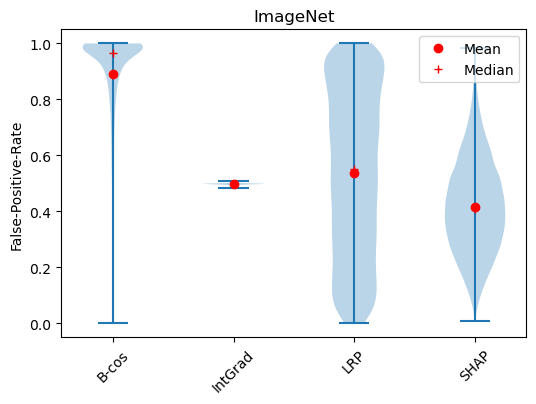}
    \caption{False-Positive-Rate}
  \end{subfigure}
  \hfill
  \begin{subfigure}[t]{0.45\linewidth}
    \includegraphics[width=\linewidth]{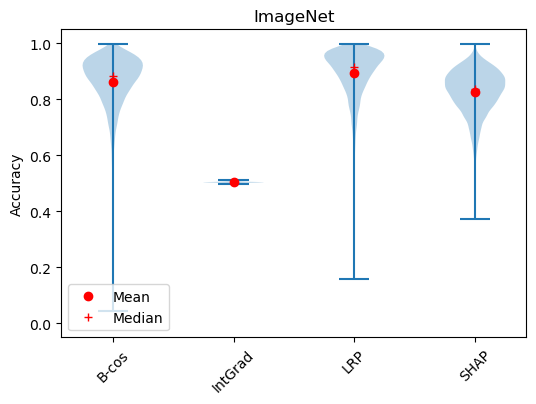}
    \caption{Accuracy}
  \end{subfigure}
  
  \caption{Results of the saliency metrics on the ImageNet mosaics for the ResNet50 model.}\label{fig:metrics-imagenet-resnet}
  \Description{Violin plots for different saliency metrics for XAI methods with negative feature importance, in the case of ResNet50 and the ImageNet dataset. The metrics each take a value between 0 and 1 on the y-axis. Precision is at the top left: IntGrad performs the worst, with a mean value of about 0.6, the values of the other methods are all between 0.8 and 0.9. On the top right is sensitivity, where B-cos and LRP have almost perfect mean values of 1.0, SHAP is at around 0.9 and IntGrad reaches approximately 0.5. Specificity is shown in the center row on the left. The worst method is B-cos with a mean value of around 0.1. The other saliency methods lie between 0.45 and 0.6, with IntGrad having a mean value of 0.5. In contrast to the large variances for LRP and SHAP, the IntGrad values show no variance. The false-negative-rate is on the right: B-cos and LRP have almost perfect mean values of 0, SHAP lies slightly above them with a mean value of approximately 0.1 and IntGrad again has a mean value of 0.5. In contrast to specificity, the variance of all values here is low. The left plot in the bottom row displays the false-positive-rate. B-cos performs the worst with a mean value of around 0.9. IntGrad and LRP lie around 0.5 with LRP having a high variance and IntGrad almost none. The mean value for SHAP is around 0.4 with a high variance. Accuracy is at the bottom right: the mean values for B-cos, LRP and SHAP lie between 0.8 and 0.9 and the mean value for IntGrad is 0.5. The variance is relatively low for all saliency methods.}
\end{figure*}

\begin{figure*}[htbp]
  \centering  
  \begin{subfigure}[t]{0.45\linewidth}
    \includegraphics[width=\linewidth]{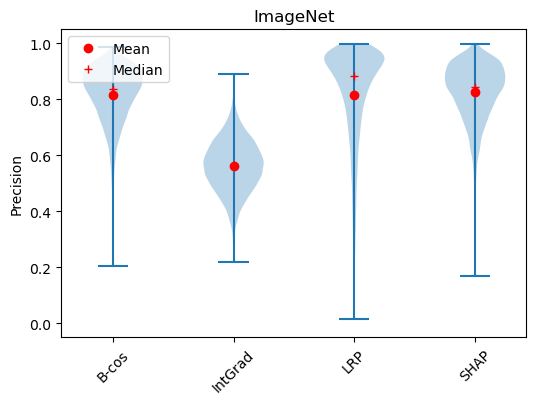}
    \caption{Precision}
  \end{subfigure}
  \hfill
  \begin{subfigure}[t]{0.45\linewidth}
    \includegraphics[width=\linewidth]{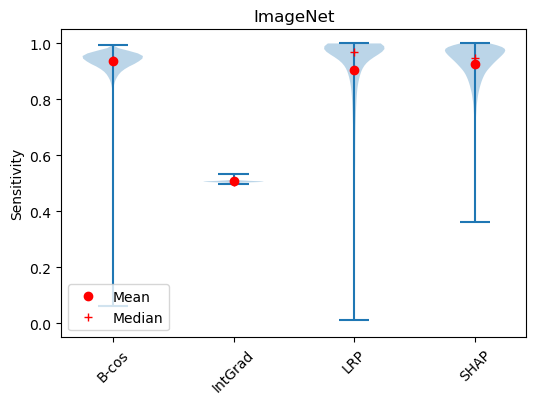}
    \caption{Sensitivity}
  \end{subfigure}

  \begin{subfigure}[t]{0.45\linewidth}
    \includegraphics[width=\linewidth]{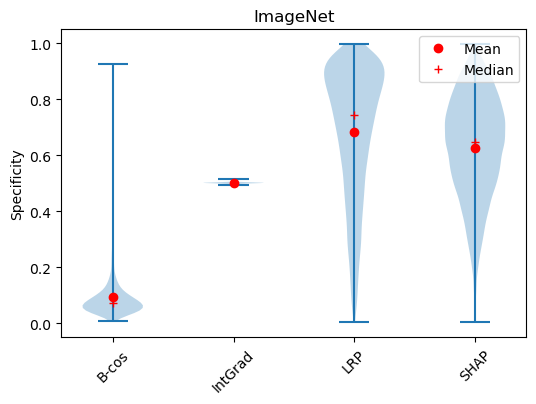}
    \caption{Specificity}
  \end{subfigure}
  \hfill
  \begin{subfigure}[t]{0.45\linewidth}
    \includegraphics[width=\linewidth]{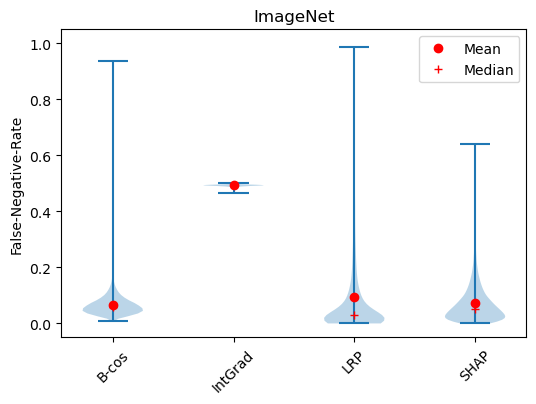}
    \caption{False-Negative-Rate}
  \end{subfigure}

  \begin{subfigure}[t]{0.45\linewidth}
    \includegraphics[width=\linewidth]{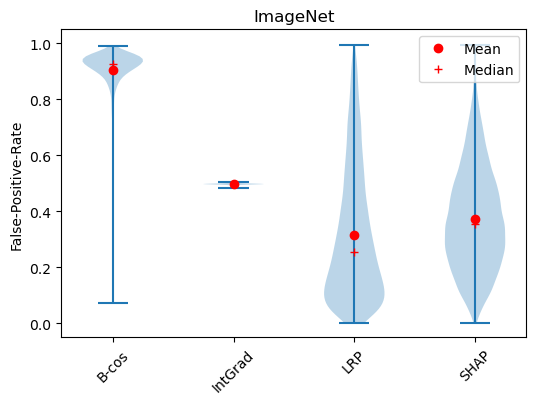}
    \caption{False-Positive-Rate}
  \end{subfigure}
  \hfill
  \begin{subfigure}[t]{0.45\linewidth}
    \includegraphics[width=\linewidth]{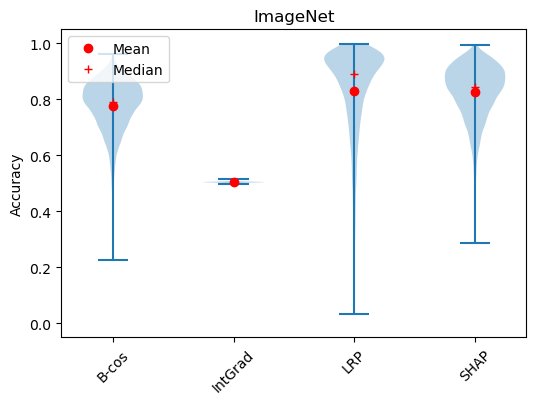}
    \caption{Accuracy}
  \end{subfigure}
  
  \caption{Results of the saliency metrics on the ImageNet mosaics for the VGG11 model.}\label{fig:metrics-imagenet-vgg}
  \Description{Violin plots for different saliency metrics for XAI methods with negative feature importance in the case of VGG11 and the ImageNet dataset. The metrics each take a value between 0 and 1 on the y-axis. Precision is at the top left: IntGrad performs the worst, with a mean value of about 0.6, the values of the other methods lie between 0.8 and 0.9. On the top right is sensitivity, where B-cos, LRP and SHAP have mean values of around 0.9 and IntGrad reaches 0.5. Specificity is shown in the center row on the left. The worst method is B-cos with a mean value of around 0.1. IntGrad has a mean value of 0.5 and LRP and SHAP lie between 0.6 and 0.7. In contrast to the high variances for LRP and SHAP, the IntGrad values show next to no variance. The false-negative-rate is on the right: B-cos, LRP and SHAP have mean values that all lie below 0.15 and IntGrad again has a mean value of 0.5. In contrast to specificity, the variance of all values here is low. The left plot in the bottom row displays the false-positive-rate. B-cos performs the worst with a mean value of around 0.9. IntGrad lies around 0.5 with almost no variance. The mean value for SHAP is around 0.4 and for LRP around 0.3. For both methods the variance is high. Accuracy is at the bottom right: the mean value for B-cos is slightly below 0.8, LRP and SHAP lie just above 0.8. The mean value for IntGrad is 0.5. The variance is relatively low for all saliency methods.}
\end{figure*}

\begin{figure*}[htbp]
  \centering
  \begin{subfigure}[t]{0.45\linewidth}
    \includegraphics[width=\linewidth]{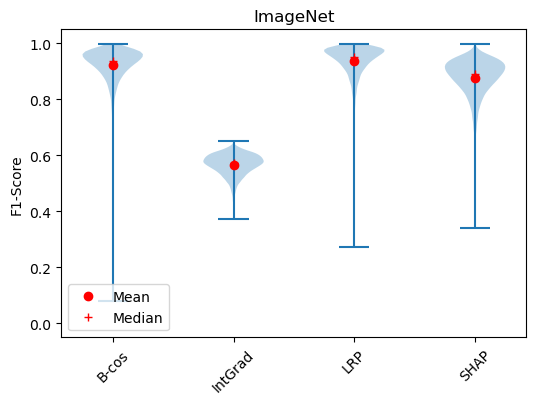}
    \caption{F1-score for ResNet50.}
  \end{subfigure}
  \hfill
  \begin{subfigure}[t]{0.45\linewidth}
    \includegraphics[width=\linewidth]{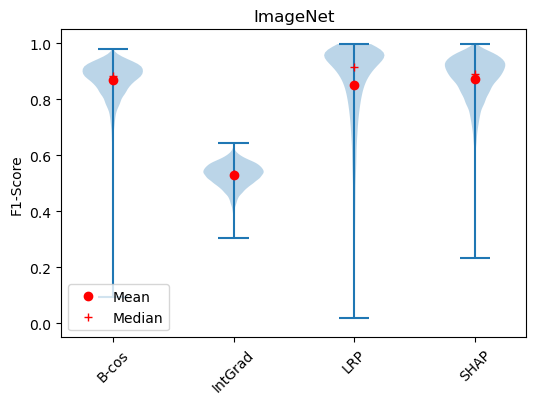}
    \caption{F1-Score for VGG11.}
  \end{subfigure}
  
  \begin{subfigure}[t]{0.45\linewidth}
    \includegraphics[width=\linewidth]{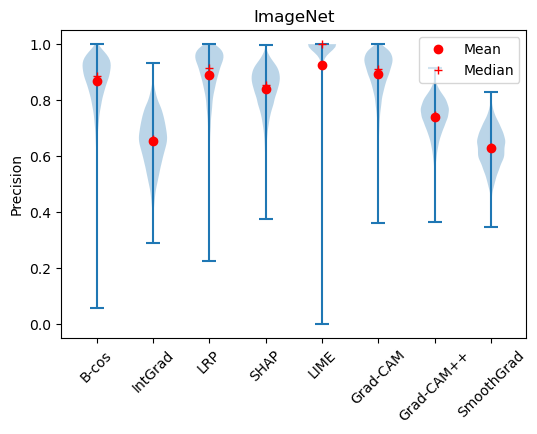}
    \caption{Precision for all methods for ResNet50.}
  \end{subfigure}
  \hfill
  \begin{subfigure}[t]{0.45\linewidth}
    \includegraphics[width=\linewidth]{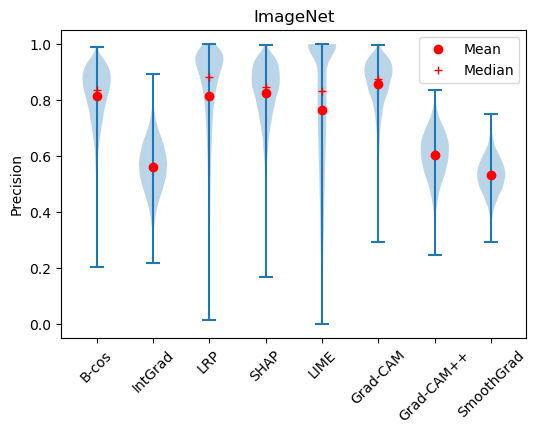}
    \caption{Precision for all methods for VGG11.}
  \end{subfigure}
  
  \caption{F1-Score and precision for both models on the ImageNet mosaics.}\label{fig:metrics-imagenet-f1-precision}
  \Description{Violin plots for the F1-score and the precision for both models on the ImageNet mosaics. The metrics each take a value between 0 and 1 on the y-axis. The F1-scores for saliency methods with negative FI are displayed in the first row, where the results for ResNet50 are on the left and the results for VGG11 are on the right. In the image on the left, the mean value of IntGrad is slightly below 0.6, that of SHAP is around 0.9 and those of B-cos and LRP are slightly higher. The results for VGG11 in the right-hand image are similar: IntGrad performs worst with a mean value of approximately 0.55. The values for B-cos, LRP and SHAP lie between 0.8 and 0.9. The variance is relatively low for all saliency methods for both models. The bottom row displays the precision for all saliency methods regarded, again with the results for ResNet50 on the left and for VGG11 on the right. On the left, the mean values of all methods lie above 0.6, with IntGrad and SmoothGrad performing worst with values around 0.65. Next is Grad-CAM++ with a mean of around 0.75 and the remaining values have mean values above 0.8. LIME is the method with the highest mean precision of approximately 0.9. On the right, the results are slightly worse: all mean values lie above 0.5, with IntGrad and SmoothGrad having the worst values around 0.55. Grad-CAM++ has a mean of around 0.6 and the remaining values are distributed around 0.8. The best performing method in this case is Grad-CAM with a mean precision of approximately 0.85.}
\end{figure*}

In the following, a discussion of the results for the saliency metrics on ImageNet can be found. Figure~\ref{fig:metrics-imagenet-resnet} shows most metrics for ResNet50 for B-cos, \ac{IntGrad}, \ac{LRP} and \ac{SHAP}, Figure~\ref{fig:metrics-imagenet-vgg} displays the same metrics for VGG11 and Figure~\ref{fig:metrics-imagenet-f1-precision} shows the F1-score for these methods and the precision for all considered saliency methods.
For both models, some similarities can be observed:
The saliency metric results of \ac{IntGrad} are close to $0.5$ with a low variance for all metrics but precision and F1-score.
Overall, the range of saliency metric values is wide (sometimes spanning from $0$ to $1$), although with the values usually concentrated on a smaller range. This can be explained by some mosaics being easier and some more difficult to explain for each of the methods, which is expected when using random images from a dataset as diverse as ImageNet.
While the B-cos models do perform well in some metrics (precision, sensitivity, false-negative-rate, accuracy and F1-score), they consistently perform bad in specificity and false-positive-rate (with values close to $0$ and $1$ respectively), showing their failure to attribute negative \ac{FI} correctly and a strong bias towards positive \ac{FI} as discussed in Section~\ref{ssec:general-findings}.
Overall, based on the median performances, \ac{LRP} and \ac{SHAP} seem to be the best methods, with \ac{SHAP} beating out \ac{LRP} regarding the variance of specificity and false-positive-rate for ResNet50, while the distribution of saliency results for those metrics is better for \ac{LRP} than for \ac{SHAP} with VGG11.
This behaviour does not show in the F1-score, which is a harmonized mean of precision and recall. But due to the higher magnitudes for positive \ac{FI}, the F1-score mainly shows how well the positive \ac{FI} is distributed. This is not surprising given that the F1-score can be rewritten to $\frac{2 tp}{2 tp + fp + fn}$.
This shows that---if correct distribution of negative \ac{FI} matters for a use-case---specificity and false-positive-rate should be considered along with one of the other metrics.

For the methods with only positive \ac{FI}, solely the precision can be calculated. 
The results in Figure~\ref{fig:metrics-imagenet-f1-precision} show that the ranking of saliency methods differs between the two models, an effect especially prominent for \ac{LIME}, which provides the best mean precision for ResNet50, but only the fifth-best for VGG11.
Based on the precision, \ac{LIME} seems to be the best-performing method for ResNet50 (although with a higher variance than the other methods) and \ac{Grad-CAM} for VGG11.
For both datasets, they are closely followed by B-cos, \ac{LRP} and \ac{SHAP}.

\subsection{Inter-rater Reliability} \label{a-ssec:inter-rater-reliability}
For detailed results for inter-rater reliability for ResNet50, see Table~\ref{tab:alpha-resnet}, for VGG11, see Table \ref{tab:alpha-vgg}.
Note that some of the used saliency methods only provide positive \ac{FI}, thus only the precision reliability can be calculated for them.
These results can be found in Table \ref{tab:alpha-positive}.
The findings for the inter-rater reliability are discussed in Section~\ref{sec:inter-rater-reliability}.

\begin{table*}[tpb]
    \centering
    \caption{Krippendorff's $\alpha$ for B-cos, \ac{IntGrad}, \ac{LRP} and \ac{SHAP} for all metrics for ResNet50.}
    \label{tab:alpha-resnet}
    \begin{tabular}{p{3cm}m{1.25cm}m{1.25cm}m{1.25cm}m{1.25cm}m{1.25cm}m{1.25cm}m{1.25cm}}
        \toprule
        & Precision & Sensitivity & False-Negative-Rate & False-Positive-Rate & Specificity & Accuracy & F1-Score \\
        \midrule
        Cars/Cats dataset & 0.81 & 0.89 & 0.89 & 0.82 & 0.82 & 0.83 & 0.85 \\
        Mountain Dogs dataset & 0.24 & 0.98 & 0.98 & 0.96 & 0.96 & 0.35 & 0.67 \\
        ImageNet & 0.61 & 0.85 & 0.85 & 0.49 & 0.49 & 0.69 & 0.74 \\
        \bottomrule
    \end{tabular}
\end{table*}

\begin{table*}[tbp]
    \centering
    \caption{Krippendorff's $\alpha$ for B-cos, \ac{IntGrad}, \ac{LRP} and \ac{SHAP} for all metrics for VGG11.}
    \label{tab:alpha-vgg}
    \begin{tabular}{p{3cm}m{1.25cm}m{1.25cm}m{1.25cm}m{1.25cm}m{1.25cm}m{1.25cm}m{1.25cm}}
        \toprule
        & Precision & Sensitivity & False-Negative-Rate & False-Positive-Rate & Specificity & Accuracy & F1-Score \\
        \midrule
        Cars/Cats dataset& 0.56 & 0.59 & 0.59 & 0.79 & 0.79 & 0.64 & 0.56 \\
        Mountain Dogs dataset & 0.15 & 0.72 & 0.72 & 0.67 & 0.67 & 0.14 & 0.52 \\
        ImageNet & 0.52 & 0.58 & 0.58 & 0.67 & 0.67 & 0.62 & 0.57 \\
        \bottomrule
    \end{tabular}
\end{table*}

\begin{table*}[tbp]
    \centering
    \caption{Krippendorff's $\alpha$ for B-cos, \ac{IntGrad}, \ac{LRP},\ac{SHAP}, \ac{LIME}, \ac{Grad-CAM}, \ac{Grad-CAM}++ and SmoothGrad for precision for ResNet50 and VGG11.}
    \label{tab:alpha-positive}
    \begin{tabular}{p{4cm}ccc}
        \toprule
        & ResNet50 & VGG11 \\
        \midrule
        Cars/Cats dataset & 0.88 & 0.56 \\
        Mountain Dogs dataset & 0.25 & 0.14 \\
        ImageNet & 0.71 & 0.56\\
        \bottomrule
    \end{tabular}
\end{table*}

\subsection{Inter-method Reliability} \label{a-ssec:inter-method-reliability}
\begin{figure*}[htbp]
  \centering

  \begin{subfigure}[t]{\textwidth}
    \centering
    \includegraphics[width=0.7\textwidth]{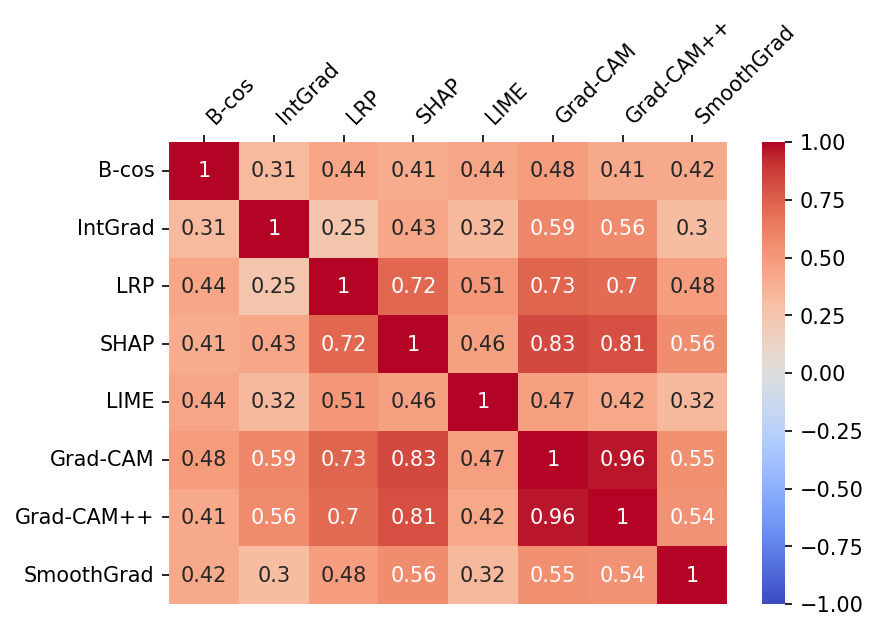}
    \caption{Spearman's $\rho$ on the Mountain Dogs dataset for the precision metric for ResNet50.}
    \label{sfig:corr-dogs}
  \end{subfigure}

  \begin{subfigure}[t]{\textwidth}
    \centering
    \includegraphics[width=0.7\textwidth]{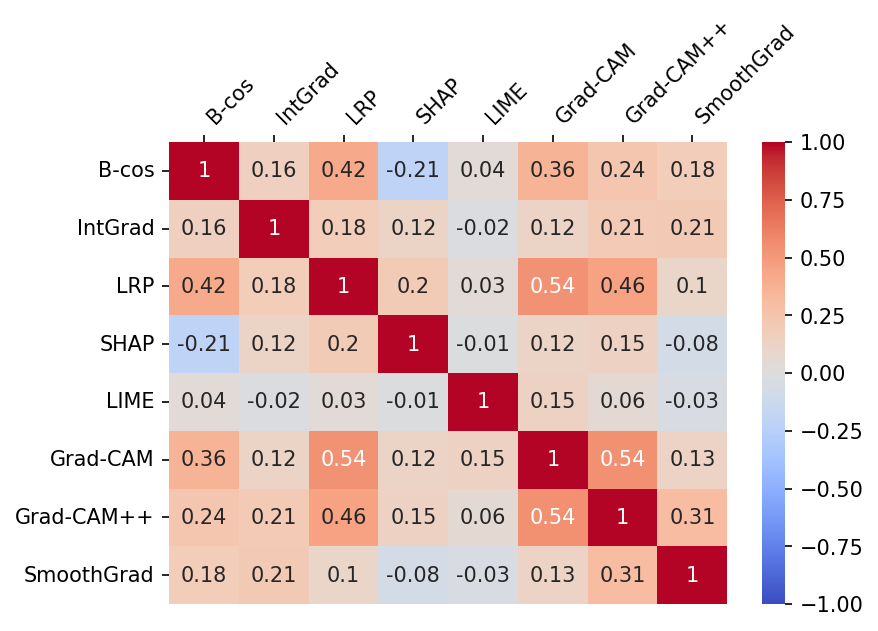}
    \caption{Spearman's $\rho$ on the Cars/Cats dataset for the precision metric for ResNet50.} 
    \label{sfig:corr-carsncats}
  \end{subfigure}
  
  \caption{Spearman's $\rho$ for the precision metric on different datasets. While the dataset with difficult to distinguish classes produces high correlation values (nearing 1 for \ac{Grad-CAM}/\ac{Grad-CAM}++), the dataset with easy to distinguish classes produces mostly random correlations, with the highest one between \ac{Grad-CAM}/\ac{Grad-CAM}++ and \ac{Grad-CAM}/\ac{LRP} with 0.54.}\label{fig:corrs}
  \Description{This figure shows Spearman's $rho$ in different colours. It can be seen that the results differ between easy and difficult to distinguish classes. For easy to distinguish classes, barely any relevant correlations can be seen with nearly all below 0.4, while for difficult to distinguish classes, most methods are correlated with barely any below 0.4.}
\end{figure*}

Detailed results for Spearman's $\rho$ correlation can be found in Figure~\ref{fig:corrs} for ResNet50 on the Cars/Cats dataset and on the Mountain Dogs dataset.
These correlation values differ between the two datasets, as for the more difficult to distinguish classes (Mountain Dogs, Figure~\ref{sfig:corr-dogs}), some methods yield highly correlated precision values, showing that these methods tend to perform similarly on the same images.
This could be due to most images being difficult to distinguish and some showing clear differences (or none at all) between the dog breeds, thus effectively producing good (bad) performances on the same images.
For the easier to distinguish classes (Cars/Cats, Figure~\ref{sfig:corr-carsncats}), no clear correlation tendencies exist, with the highest value below $0.6$ and most being close to 0, with some even below 0.
This shows that the saliency methods do not consistently agree on which of the mosaics in this dataset is easier or more difficult to attribute correctly.
Since this dataset approximates real-world use-cases better than the Mountain Dogs dataset, it can be concluded that for real-world applications, saliency methods will likely struggle with different images and the difficulty of explaining a decision is not inherent to images but related to the used saliency method.
In Section~\ref{sec:results}, this was summarized as ``\emph{The saliency methods tend to work individually but some of them fail jointly}''.

\end{document}